\documentclass[12pt,draftclsnofoot,onecolumn]{IEEEtran}

\usepackage[ruled]{algorithm2e}
\usepackage{stfloats}
\usepackage{bbm}
\usepackage{amssymb}
\usepackage{amsmath,amsthm,amssymb}
\usepackage{graphicx}
\usepackage{subfigure}
\usepackage{booktabs} 
\usepackage{epsfig,epsf,color,balance,cite}
\usepackage{threeparttable}
\usepackage{verbatim}
\usepackage{url}
\usepackage{booktabs}
\usepackage{bm}
\usepackage{caption}
\usepackage{diagbox}
\usepackage{multirow}
\usepackage{enumerate}
\usepackage{pifont}
\usepackage{tablefootnote}
\usepackage{cite}
\usepackage{tabularx}

\newcommand{\rev}{\textcolor{black}}

\begin{document}

\captionsetup[figure]{name={Fig.},labelsep=period}
%

\title{Task-Oriented Low-Label Semantic Communication With Self-Supervised Learning}
\author{Run Gu,~Wei Xu,~\IEEEmembership{Fellow,~IEEE},~Zhaohui Yang,~\IEEEmembership{Member,~IEEE},\\~Dusit Niyato,~\IEEEmembership{Fellow,~IEEE}, and Aylin Yener,~\IEEEmembership{Fellow,~IEEE}
	
%
%
%

	\thanks{Part of this work was presented in \textit{WOCC 2024} \cite{gu2024self}.}
	\thanks{Run Gu and Wei Xu are with National Mobile Communications Research Laboratory, Southeast University, Nanjing 210096, China, and also with Purple Mountain Laboratories, Nanjing 211111, China (e-mail: \{rung, wxu\}@seu.edu.cn).}
	\thanks{Zhaohui Yang is with the Zhejiang Lab, Hangzhou 311121, China, and also with the College of Information Science and Electronic Engineering, Zhejiang University, Hangzhou, Zhejiang 310027, China (yang\_zhaohui@zju.edu.cn).}
	\thanks{Dusit Niyato is with the School of Computer Science and Engineering, Nanyang Technological University, Singapore 308232 (dniyato@ntu.edu.sg).}
	\thanks{Aylin Yener is with the Department of Electrical and Computer Engineering, The Ohio State University, OH 43210, USA (yener@ece.osu.edu).}
	}

\maketitle
\begin{abstract}

Task-oriented semantic communication enhances transmission efficiency by conveying semantic information rather than exact messages. Deep learning (DL)-based semantic communication can effectively cultivate the essential semantic knowledge for semantic extraction, transmission, and interpretation by leveraging massive labeled samples for downstream task training. 
In this paper, we propose a self-supervised learning-based semantic communication framework (SLSCom) to enhance task inference performance, particularly in scenarios with limited access to labeled samples. Specifically, we develop a task-relevant semantic encoder using unlabeled samples, which can be collected by devices in real-world edge networks. 
To facilitate task-relevant semantic extraction, we introduce self-supervision for learning contrastive features and formulate the information bottleneck (IB) problem to balance the tradeoff between the informativeness of the extracted features and task inference performance. Given the computational challenges of the IB problem, we devise a practical and effective solution by employing self-supervised classification and reconstruction pretext tasks. We further propose efficient joint training methods to enhance end-to-end inference accuracy over wireless channels, even with few labeled samples. We evaluate the proposed framework on image classification tasks over multipath wireless channels. Extensive simulation results demonstrate that SLSCom significantly outperforms conventional digital coding methods and existing DL-based approaches across varying labeled data set sizes and SNR conditions, even when the unlabeled samples are irrelevant to the downstream tasks.      
 
 

\end{abstract}


%
\IEEEpeerreviewmaketitle

\section{Introduction}	
	\IEEEPARstart{W}{ith} the widespread deployment of edge devices and the rapid development of artificial intelligence (AI), an impressive landscape of connected intelligence is emerging\cite{letaief2021edge,xu2023toward,yao2025byzantine,shi2025combating}.
	In a network with edge intelligence, devices transmit data to a server wirelessly at the edge, and the received data can be analyzed using AI to support various applications\cite{guo2022distributed,yao2024wireless}.
	Most such applications experience a surging demand for massive data transmission\cite{xu2023reconfiguring}, such as high-resolution images for surveillance. This unprecedented demand on bandwidth poses a challenge for conventional communication systems. Semantic communication is a novel paradigm for efficiently handling such high-volume data transmissions\cite{gunduz2022beyond,guler2014compressing,guler2014semantic}. 
	
	By transmitting semantic information specific to certain tasks within large volumes of raw data, semantic communication holds the potential to enhance communication efficiency, meeting the requirements of next-generation applications~\cite{huang2024stacked}. It is typically challenging to effectively extract and interpret semantic information for a variety of tasks using traditional mathematical models~\cite{qin2022semantic}.
	As such AI-aided semantic extraction has garnered significant attention. One recent line of work capitalizes on pre-trained models for efficient conveying of semantic information~\cite{kutay2023semantic,kutay2024classification}. Another notable approach focuses on end-to-end deep learning (DL)-aided semantic communications. 
	DL has exhibited remarkable efficacy in feature extraction, leading to widespread applications in wireless communication
	~\cite{ye2020deep,chaccour2022less}. In DL-based semantic communication systems, joint source-and-channel coding (JSCC) methods ensures reliable data recovery, collectively enhancing end-to-end performance compared to conventional schemes. Semantic codecs, implemented via DL networks, cultivate and share semantic knowledge among transceivers, fostering the extraction, transmission, and interpretation of semantic information~\cite{han2022semantic}. These studies are broadly categorized into two objectives, i.e., data reconstruction and task inference.
	
	For data reconstruction, a common approach for the DL network training involves an autoencoder algorithm, where semantic knowledge is acquired from the source data, enabling the transmitted semantic information to be robust against channel distortion\cite{bourtsoulatze2019deep,xie2021deep}. In the context of task inference, DL networks are typically trained using labeled data, which combines label information with source data. Supervised with the labeled data, semantic knowledge relevant to specific tasks is captured from the source data. Most existing task-oriented semantic communication studies utilize supervised learning algorithms for network training, implemented through single-phase\cite{shao2021learning,zhang2022deep,huang2024stacked} or multi-phase frameworks~\cite{xie2022task,jankowski2020wireless}.
	  
	
	These task-oriented communication studies usually rely on large volumes of labeled samples, often requiring tens of thousands. However, in wireless communication applications, collecting task-specific labeled samples presents significant challenges due to various dynamics in complex wireless environments, along with expensive and labor-intensive annotation process\cite{qin2019deep,he2019model}. \rev{These difficulties are common in real-world applications such as target sensing~\cite{mateos2024semi} and disaster monitoring~\cite{liu2024stacked} in integrated sensing and communication, traffic volume prediction and vehicle monitoring in traffic management\cite{rabby2019review}, and equipment fault detection in Industrial IoT\cite{wu2022survey}}. The resulting scarcity of labeled samples hinders effective cultivation of semantic knowledge, thereby compromising the performance of transmitted semantic information over wireless channels in terms of task inference. Therefore, it is regarded a crucial challenge to enhance task inference performance in wireless semantic communication systems with limited labeled samples.
	
		
	Significant efforts have been made to enhance inference performance without sufficient labeled samples in the field of computer vision. These approaches can be broadly categorized into three groups, i.e., methods based on data diversity, methods involving cross-domain transfer, and methods exploiting unlabeled samples. 
	
	The first category of methods based on data diversity focuses on increasing the variety of data through augmentation techniques\cite{khosla2020enhancing} and generative models\cite{shivashankar2023semantic}. These techniques depend on the realism of the augmentations and the quality of the synthetic data generated since augmented or synthetic data might fail to capture the task-relevant characteristics and the effect of wireless environments. The second category of methods involving cross-domain transfer, including transfer learning\cite{pan2010survey}, meta-learning\cite{finn2017model}, and domain adaptation\cite{song2018improving}, seeks to leverage knowledge across different tasks or domains.
	However, the effectiveness of cross-domain transfer is highly dependent on the similarity in tasks or datasets, which is often difficult to achieve, especially in  heterogeneous wireless communication applications. Since unlabeled samples offer a wealth of information beyond what artificial annotations can provide, methods exploiting unlabeled samples have emerged as a promising alternative. Common unsupervised learning algorithms, such as principal component analysis, clustering, autoencoders, and generative models are not well-suited for task-oriented semantic communication, as they lack the ability to prioritize task-relevant features. Specifically, principal component analysis and clustering often capture general-purpose features rather than those directly tied to specific tasks. Meanwhile, autoencoders and generative models focus on data reconstruction and introduce additional computation complexity through their decoder modules which are not required for task inference purposes.

	Self-supervised learning is a celebrated unsupervised learning algorithm that considers intrinsic data features and learning objectives to construct self-supervision and pretext tasks\cite{han2021pre}. By optimizing the pretext tasks within self-supervised frameworks, we can extract the intended features from input data, such as contextual features\cite{pathak2016context} derived from generative schemes and contrastive features\cite{wu2018unsupervised} based on discriminative schemes. It is worth mentioning that discriminative schemes, as used in contrastive learning, excel at isolating task-relevant information from irrelevant data by clustering semantically similar features while repelling dissimilar ones\cite{khan2022contrastive}. This approach has also demonstrated strong generalizability across various tasks\cite{taghanaki2022self}. It therefore emerges as a well-suited solution for task-oriented communication systems. 
	Specifically for applications of semantic communication, the study in \cite{chaccour2022disentangling} adopted two discrimination tasks \cite{chen2020simple} to extract contrastive features, effectively reducing the transmission length of data points. Moreover, the features learned through self-supervised methods are commonly fine-tuned for downstream tasks, a process referred to as semi-supervised learning\cite{han2021pre}. This approach leverages both labeled and unlabeled samples, where the labeled samples serve as a foundation for associating the learned features with task-specific objectives. In \cite{tang2023contrastive}, the learned image contrastive features through a discrimination task were fine-tuned using labeled samples to serve both transmission and classification tasks. However, existing approaches to utilizing unlabeled samples exhibit a data-oriented focus rather than a task-oriented one, even in \cite{tang2023contrastive} where the classification task was conducted using reconstructed images.
	
	To facilitate efficient task-oriented communication, the information bottleneck (IB) principle \cite{tishby2000information} was applied in \cite{shao2021learning} to extract the minimal yet sufficient features for transmission, which effectively reduces communication overhead over noisy channels. However, as previously mentioned, this study utilized massive labeled samples for network training, but the scarcity of labeled samples may hinder its overall effectiveness. This paper aims to enable efficient network learning for robust task inference performance of intelligent edge applications, particularly in scenarios where only limited portions of samples are task-specific and annotated. The main contributions of this paper are summarized as follows.
	\vspace{-5pt}
\begin{itemize}
	\item As a nontrivial extension of existing task-oriented semantic communication works, we develop an efficient learning approach that leverages both labeled and unlabeled samples to enhance task inference performance. The proposed approach uniquely utilizes task-specific unlabeled samples for training the semantic extraction module. Even with a small set of labeled data and when the unlabeled samples are irrelevant to downstream tasks, this approach facilitates effective semantic transmission and interpretation, especially under multipath fading channels.
	\item To extract task-relevant semantics for efficient communication, we develop a semantic encoder network by incorporating the information bottleneck (IB) principle and harnessing contrastive features learned from unlabeled data. Specifically, we introduce the self-supervision on unlabeled samples to connect the contrastive features to downstream tasks. To further ensure that the contrastive features, serving as semantic information, are minimal yet sufficient for these tasks,  we formulate the IB problem that balances the tradeoff between the informativeness of semantic information and task inference performance. Then, a practical and effective solution is devised to address the IB problem via classification and reconstruction pretext tasks. The resulting semantic information effectively retains inference precision and also exhibits strong generalization to unseen data.
	\item We demonstrate the effectiveness of the proposed approach for jointly training semantic extraction, transmission, and interpretation modules. This ensures satisfactory end-to-end inference performance while compressing the extracted semantic information for enhanced communication efficiency. The resulting JSCC network effectively mitigates the distortion, especially under multhpath wireless fading channels.
	\item We conduct extensive simulations across different labeled data set sizes and SNR conditions using two network structures, each tailored for performance-oriented and resource-limited applications. The results show that the proposed approach outperforms digital coding schemes, training from scratch, and transferring pre-trained model in classification task-oriented communication. The performance gains, driven by the learned semantic encoder and JSCC networks, are clearly identified, with the improvements being particularly significant in challenging scenarios, such as low SNR conditions and few labeled samples. Moreover, we validate the generalizability of the proposed approach under complex impairments including data distribution shifts and partial label missing. Ablation studies further highlight the positive impact of the reconstruction pretext task on classification accuracy and demonstrate the necessity of the auxiliary projection layer in these challenging scenarios.
	\end{itemize} 
	
	The rest of this paper is organized as follows. Section \uppercase\expandafter{\romannumeral2} introduces the system model and presents design objectives of task-oriented semantic communication. Section \uppercase\expandafter{\romannumeral3} details the two-stage learning framework and network structure of the proposed SLSCom. Simulation results are presented in Section \uppercase\expandafter{\romannumeral4}, and conclusions are drawn in Section \uppercase\expandafter{\romannumeral5}.
		
	\textit{Notations}: Random variables and their realizations are represented by normal-face lower and uppercase letters, respectively. Boldface lower and uppercase letters denote column vectors and matrices, respectively. $\mathbb{R}^{n\times m}$ and $\mathbb{C}^{n\times m}$ represent the sets of real and complex matrices of size $n\times m$, respectively. The entropy of $X$, the mutual information between $X$ and $Y$, and the entropy of $Y$ conditioned on $X$ are respectively denoted by $H(X)$, $I(X;Y)$, and $H(Y|X)$. The operator $\|\cdot \|_2$ takes the Euclidean norm. Let the $\cdot$ operation denote the inner product, i.e., $\mathbf{x}\cdot \mathbf{y} = \mathbf{x}^\text{T}\mathbf{y}$. The Gaussian distribution with mean $\mu$ and covariance $\sigma^2$ is $\mathcal{CN}(\mu,\sigma^2)$. 
	
\section{System Model and Problem Description}\label{sec2}
	\begin{figure*}[t]
		\centering
		\includegraphics[width=\linewidth]{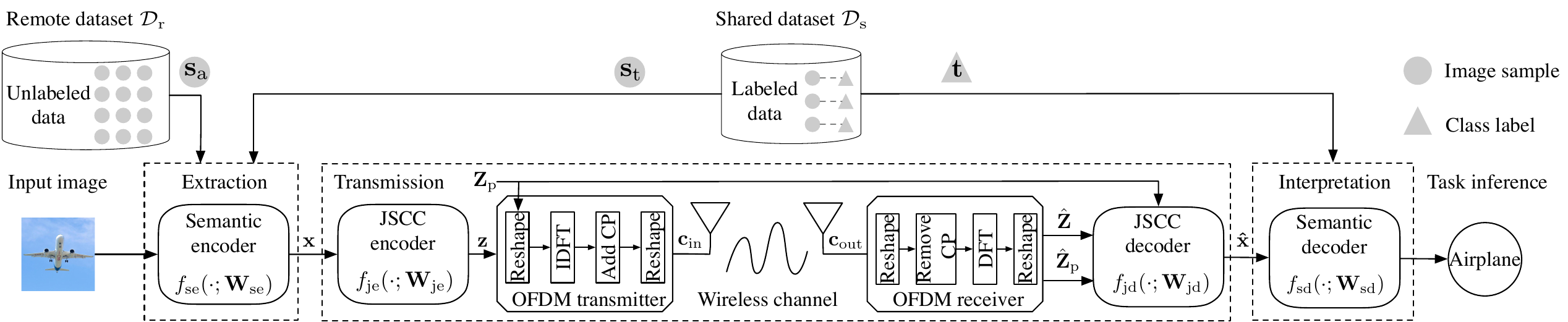}
		\caption{\rev{Diagram of the proposed semantic communication.}}
		\label{commu model}
	\end{figure*}
	
	We consider a task-oriented semantic communication system for intelligent visual applications where limited portions of samples are task-specific and annotated. As depicted in Fig.~\ref{commu model}, 
	both the device and the server are equipped with neural networks (NNs). These NNs are categorized into three components: a semantic encoder network, a JSCC network, and a semantic decoder network. They are responsible for semantic extraction, transmission, and interpretation, respectively. There are two types of datasets available for training NN, characterizing the low-label scenarios. A dataset on the $\textit{remote}$ device side, denoted by $\mathcal{D}_\text{r}\triangleq\{(\mathbf{s}^{(m)}_\text{a})^{M_1}_{m=1}\}$, comprises data alone $\bf s$. Another dataset $\textit{shared}$ among transceivers, denoted by $\mathcal{D}_\text{s}\triangleq\{(\mathbf{s}_\text{t}^{(m)},\mathbf{t}^{(m)})^{M_2}_{m=1}\}$, contains a limited number of labeled samples. Here, $\mathbf{s}_\text{t}$ and $\bf t$ represent samples paired with their respective labels for the downstream task. Particularly, human-annotated labels are regarded as a compact kind of task-relevant information. 
		
	\subsection{Task-oriented Semantic Communication}\label{2a}
	The communication process is as follows. At the transmitter, the device captures an image, denoted by a reshaped vector $\mathbf{s} \in \mathbb{R}^{L_\text{i}}$ with length $L_\text{i}$, as the input of the semantic encoder network. The semantic information, denoted by $\mathbf{x} \in \mathbb{R}^{L_\text{s}}$, with length $L_\text{s}$ is extracted by 
	\begin{align}\label{feat}
		\mathbf{x}=f_{\text{se}}(\mathbf{s} ;{\mathbf{W}}_{\text{se}}),
	\end{align}
	where $f_{\text{se}}(\cdot;\mathbf{W}_{\text{se}})$ represents the semantic encoder network with a trainable parameter set $\mathbf{W}_{\text{se}}$. Besides, we use the compression ratio (CR) of $\mathbf{x}$, denoted by $\text{CR}_\text{s}$, to characterize the source compression ratio. The calculation of $\text{CR}_\text{s}$ is given by $\text{CR}_\text{s} \triangleq L_\text{s}/L_\text{i}$.
	
	Due to limited communication resource and complex wireless conditions, the semantic information is subsequently encoded by a JSCC encoder network, which is expressed as
	\begin{align}\label{eq:encoded}
		\mathbf{z}=f_{\text{je}}(\mathbf{x};\mathbf{W}_{\text{je}}),
	\end{align}
	where $\mathbf{z} \in \mathbb{C}^{L_\text{c}}$ is the mapped complex-valued vector with length $L_\text{c}$ and $f_{\text{je}}(\cdot;\mathbf{W}_{\text{je}})$ represents the JSCC encoder network with a trainable parameter set $\mathbf{W}_{\text{je}}$. The encoded symbol $\bf z$ is normalized to guarantee the average power constraint, i.e., $\lVert \mathbf{z}\rVert^2_2 /{L_\text{c}} = 1$. Similar to $\text{CR}_\text{s}$, we denote $\text{CR}_\text{c}$ as the CR of $\bf z$, representing the channel-utilization per pixel and defined by $\text{CR}_\text{c} \triangleq L_\text{c}/L_\text{i}$.
	
	Consider that the device is connected to the server via a wireless fading channel, which is usually modeled as 
	\begin{align}\label{mpc}
		\mathbf{c}_{\text{out}}=\mathbf{h} \ast \mathbf{c}_{\text{in}} + \mathbf{n},
	\end{align}
	where $\mathbf{h} \in \mathbb{C}^{L_\text{h}}$ is the multipath channel of $L_\text{h}$ paths, $\mathbf{c}_{\text{in}}$ is the channel input vector, $\mathbf{c}_{\text{out}}$ is the channel output vector, the operator $\ast$ represents a convolution operation, and $\mathbf{n}$ is the noise with energy $\sigma^2_\text{n}$.
	
	To mitigate inter-symbol interference in the multipath fading channels, we adopt orthogonal frequency division multiplexing (OFDM) with $K$ subcarriers, comprising $N_\text{s}$ OFDM symbols for data and $N_\text{p}$ OFDM symbols for pilots. \rev{Specifically, we reshape $\bf z$ to data symbol $\mathbf{Z} \in \mathbb{C}^{K \times N_\text{s}}$. Pilot symbol $\mathbf{Z}_\text{p}$ is then inserted into $\mathbf{Z}$ along the OFDM symbol dimension. As seen in Fig.~\ref{commu model}, after performing inverse discrete Fourier transform (IDFT) and adding cyclic prefix (CP) with a length of $L_\text{cp}$ in the OFDM transmitter, the resulting time-domain signal is reshaped into vector $\mathbf{c}_{\rm in}$.} Here, $KN_\text{s}=nL_\text{c}$, where $n=1,2,\ldots$, signifying a single OFDM frame transmits information of $n$ images. After the inverse operations of CP removal and DFT applied to $\mathbf{c}_\text{out}$ by the OFDM receiver at the server, the equivalent frequency-domain transmission is expressed by
	\begin{equation}\label{channel}
		\begin{split}
	\mathbf{\hat{Z}}[k,i]&=\mathbf{H}[k] \mathbf{Z}[k,i]+\mathbf{N}[k,i],~  i=1,\dots,N_\text{s},  \\
	\mathbf{\hat{Z}}_\text{p}[k,j]&=\mathbf{H}[k] \mathbf{Z}_\text{p}[k,j]+\mathbf{N}_\text{p}[k,j], ~j=1,\dots,N_\text{p},
		\end{split}
	\end{equation}
	where $\mathbf{H}[k] $ is frequency response of the channel $\bf h$ at the $k$th subcarrier, and $\bf \hat{Z}$ and $\mathbf{\hat{Z}}_\text{p}$ represent the received data and pilot symbols, respectively. Both $\bf N$ and $\mathbf{N}_\text{p}$ denote the noise samples. The average signal-to-noise ratio (SNR) over subcarriers can be expressed as
	\begin{align}\label{eq:snr}
		\text{SNR}=\frac{\sum^K_{k=1}\sum^{N_\text{s}}_{i=1} \lvert \mathbf{H}[k] \mathbf{Z}[k,i]\rvert^2}{KN_\text{s}\sigma^2_\text{n}}.
	\end{align}

	Given $\mathbf{Z}_\text{p}$, $\mathbf{\hat{Z}}$, and $\mathbf{\hat{Z}}_\text{p}$, a JSCC decoder is utilized to decode the semantic information $\mathbf{\hat{x}}\in \mathbb{R}^{L_\text{c}}$, which is represented as
	\begin{align}\label{eq:decoded}
		\mathbf{\hat{x}} = f_{\text{jd}}(\mathbf{\hat{Z}},\mathbf{Z}_\text{p}, \mathbf{\hat{Z}}_\text{p};\mathbf{W}_{\text{jd}}),
	\end{align}
	where $f_\text{jd}(\cdot;\mathbf{W}_\text{jd})$ represents the JSCC decoder network with a trainable parameter set $\mathbf{W}_\text{jd}$.
	
	For a specific task, the recovered semantic information is interpreted to perform the final task, as shown in Fig.~\ref{commu model}. It is expressed as
	\begin{align}\label{eq:inference}
		\mathbf{y} = f_\text{sd}(\mathbf{\hat{x}};\mathbf{W}_\text{sd}),
	\end{align}
	where $\bf y$ is the task inference and $f_\text{sd}(\cdot;\mathbf{W}_\text{sd})$ is the semantic decoder network with a trainable parameter set $\mathbf{W}_\text{sd}$.  
		
	\subsection{Problem Description}

 	
 	The goal of the considered system is to achieve accurate task inference from the received semantic information. This is typically done via end-to-end learning, using a task-specific training set, i.e., $\{(\mathbf{s}^{(m)}_\text{t},\mathbf{t}^{(m)})^{M_2}_{m=1}\}$. To ensure robust task inference performance under limited communication resource, the IB principle has been further applied to transmit only task-relevant information. Inspired by the information-theoretic measures employed in\cite{shao2021learning,tsai2020self,xu2024disentangled,xu2023edge,zhang2024dibad}, we formulate the IB-based optimization problem for the aforementioned communication process, expressed as
 	\rev{\begin{equation}\label{tradeoff}
		\begin{split}
			(\text{P1})~\mathop \text{maximize}_{\mathbf{W}_\text{se},\mathbf{W}_\text{je},\mathbf{W}_\text{jd}} \ & \ -\lambda H(\hat{X}|T)+I(\hat{X};T),
		\end{split}
	\end{equation}where $\lambda$ is a positive weight balancing the two metrics, $\hat X$ and $T$ are the random variables of the received semantic information $\hat{\bf x}$ and the label $\bf t$, respectively.} The optimal parameter sets of these NNs are denoted by $\mathbf{W}^\ast_\text{se}$, $\mathbf{W}^\ast_\text{je}$, and $\mathbf{W}^\ast_\text{jd}$, respectively. 
		
	This IB-based optimization problem differs from the one in \cite{shao2021learning} in the following aspects. Firstly, existing task-oriented semantic communication works, including~\cite{shao2021learning}, typically rely on a large amount of labeled training samples, often on the order of tens of thousands, denoted by $M$. \rev{In contrast, we consider a scenario with limited availability of labeled samples, such that $M_2\ll M$.} This introduces an easily neglected constraint on the quantity of labeled samples. Additionally, we exploit the objective function using the entropy of $\hat{\bf x}$ conditioned on $\bf t$ and the mutual information between $\hat{\bf x}$ and $\bf t$ to characterize the task-irrelevant informativeness in $\hat{\bf x}$ and the inference accuracy, respectively.  
	
	We specifically clarify these metrics. The mutual information $I(\hat{X};T)$ represents the information in $\hat{X}$ that is related to $T$. Maximizing $I(\hat{X};T)$ compels $\hat{\bf x}$ to contain as much task-relevant information as possible for predicting the label $\bf t$. On the other hand, the conditional entropy $H(\hat{X}|T)$ measures the amount of information in $\hat{X}$ that is irrelevant to $T$. Minimizing $H(\hat{X}|T)$ ensures discarding of task-irrelevant information. 

	However, it is intractable to obtain these optimal parameter sets in most practical systems with few task-specific labeled samples. The sample scarcity makes it difficult to obtain the comprehensive and robust semantic knowledge, resulting in model overfitting and, consequently, poor generalizability. 

\begin{figure*}[t]
	\centering
	\includegraphics[width=\textwidth]{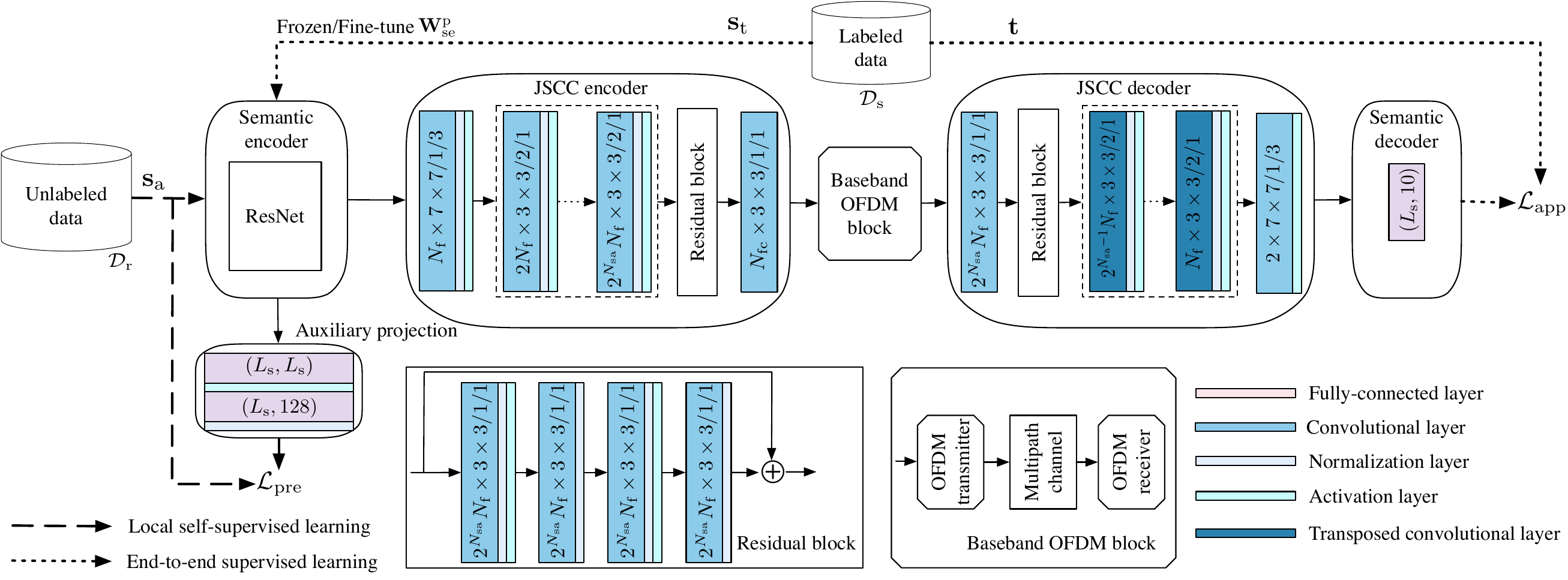}
	\caption{The framework and network overview of SLSCom. The parameters of fully-connected (FC) layers are denoted in the format of (input size, output size). The parameters of (transposed) convolutional layers are represented in the format of filters/stride/padding. }
		\label{slscArchitecture}
	\vspace{-0.4cm}
\end{figure*}
	
\section{Self-supervised Learning for Task-oriented Low-label Semantic Communication}\label{sec3}

	In this section, we propose a self-supervised learning framework for task-oriented semantic communication (SLSCom) to improve the performance for inference in wireless channels with a limited number of labeled samples. 
	The overall framework of SLSCom with NN structures is shown in Fig.~\ref{slscArchitecture}.

 \subsection{Architecture of SLSCom}
 	The self-supervisedly learned features from unlabeled data can perform well on downstream tasks, which is supported by the indisputable fact that unlabeled data inherently contains task-relevant information. Through continuous image collection by the device in practice, the remote dataset $\mathcal{D}_\text{r}$ with large-scale data can be established. \rev{We assume $M_1\leq M$, aligning with the assumption in the system model that unlabeled samples are abundant compared to the labeled ones.} 
 	
 	Based on these observations, the proposed learning framework advocates for the pre-training of the semantic encoder network $f_\text{se}$ exclusively using unlabeled data $\mathbf{s}_\text{a}$ in $\mathcal{D}_\text{r}$, without requiring external communication with a server. \rev{The decision to train only the semantic encoder, rather than jointly optimizing all encoder and decoder modules, is motivated by acquiring design flexibility and practicality in semantic communication systems. First, this approach can be seamlessly extended to semantic communication systems based on common digital transmission infrastructures, where the extracted semantic information is encoded and modulated for transmission. Second, joint training of all encoder and decoder networks typically ties their performance to specific channel conditions during training, which however requires retraining of this pre-training when conditions change. In contrast, training the semantic encoder solely at the transmitter decouples semantic extraction from channel-specific features, enabling a more generalizable semantic encoder. In real-world deployments, devices hold promise for incrementally updating the semantic encoder using locally collected task-specific data,  allowing continuous model improvement without requiring large-scale labeled datasets. The efficient, incremental updates significantly enhance the system’s practical viability, making it both scalable and efficient in dynamic wireless communication systems~\cite{chaccour2022less}.}  	
 	Nevertheless, we emphasis that joint optimization of the semantic encoder with other modules is necessary to ensure that the extracted semantic information transmitted over wireless channels excels in downstream tasks. Therefore, the proposed learning framework consists of a local self-supervised learning stage and an end-to-end supervised learning stage.
 	 
 	In the local self-supervised learning stage, our focus is learning task-relevant semantic extraction from unlabeled data in $\mathcal{D}_\text{r}$, as depicted in Fig.~\ref{slscArchitecture}. To accomplish this, we introduce self-supervision based on the popular discriminative schemes to reformulate the problem (P1), and develop pretext tasks to provide a practical and effective solution via a learning objective $\mathcal{L}_\text{pre}$.
 	During the end-to-end supervised learning stage, NNs are jointly trained using labeled samples in the shared dataset $\mathcal{D}_\text{s}$, with a learning objective $\mathcal{L}_\text{app}$. We note that the pre-trained semantic encoder parameter set $\mathbf{W}^\text{p}_\text{se}$ can either be frozen or fine-tuned. 
 	 	

	The detailed structures of the NNs is depicted in Fig.~\ref{slscArchitecture}. The proposed learning framework is applicable to various visual tasks, e.g., classification, detection, and segmentation, and is not confined to specific NN structures. Given the effectiveness of ResNet for vision tasks\cite{he2016deep}, we adopt it as the backbone for $f_\text{se}$.
	The depth of ResNet directly determines the dimensionality of the semantic information, i.e., $L_\text{s}$.
	Additionally, an auxiliary projection involving multiple fully-connected (FC) layers as well as a normalization layer is used to facilitate the learning of $f_\text{se}$ in the local self-supervised learning stage. The auxiliary projection maps the semantic information into a lower-dimensional space, aiming to eliminate prior information introduced during the establishment of self-supervision\cite{chen2020simple}, such as data transformations in the self-supervised data construction as discussed below. 
	The structure of the JSCC network follows the design principles of the commonly used convolutional networks in a row\cite{yang2022ofdm}. Specifically, the JSCC encoder network employs composite convolutions and a residual block for downsampling. 
	The corresponding JSCC decoder network follows a similar structure but in a reverse order with appropriate modifications for upsampling. The popular batch normalization and rectified linear unit (ReLU) activation function are used. 
	The structure of the semantic decoder is optimized for the specific task, such as using an FC layer for a ten-class image classification task.
	
	The network parameters, including ResNet depth, initial filter count $N_\text{f}$, channel input filter count $N_\text{fc}$, and sampling factor $N_\text{sa}$, are adaptable to diverse application scenarios. We use the CR pairs, comprising $\text{CR}_\text{s}$ of semantic information and $\text{CR}_\text{c}$ of channel input symbols, to refer to the distinct NN structures. The NN structure with larger CR values is chosen for performance-oriented applications, while the one with smaller CR values is preferred for resource-limited applications. Values of these network parameters are specified in Section \uppercase\expandafter{\romannumeral4}.
 	
\subsection{Problem Reformulation via Self-supervision}
	
 	In the self-supervised learning, self-supervision is essential as it provides data itself with pseudo-labels. Our self-supervision follows the discriminative schemes for extracting contrastive features. To relate the contrastive features with downstream tasks, we adopt a common assumption from the field of multi-view learning\cite{sridharan2008information}. Furthermore, we reformulate (P1) by incorporating the self-supervision, facilitating the learning of contrastive features guided by the IB principle.
 	
 	We start by the discriminative schemes to understand self-supervision, where each unlabeled sample $\mathbf{s}^{(m)}_\text{a}$ in the remote dataset $\mathcal{D}_\text{r}$ belongs to a distinct class indexed by $m$ and only its corresponding self-supervised data shares the same class. \rev{Let $f_\text{dc}$ denote a data construction block  to generate the self-supervised data $\mathbf{r}^{(m)}$, i.e., $\mathbf{r}^{(m)}=f_\text{dc}(\mathbf{s}^{(m)}_\text{a})$.} The class relation is expressed by
	\begin{equation}\label{eq:instdisc}
		\begin{split}
		  \mathcal{P} &= \left\{(\mathbf{s}^{(m)}_\text{a}, \mathbf{r}^{(m)}) |~ m=1,\ldots,M_1 \right\},\\ 
		  \mathcal{N} &= \left\{(\mathbf{s}^{(i)}_\text{a},\mathbf{r}^{(j)})|~\forall i,j \in \{1,\ldots,M_1\} \ \text{and} \ i\neq j\right\}.
		\end{split}
	\end{equation}
	Here, $\mathcal{P}$ represents the set of positive sample pairs, consisting of a source sample and its corresponding self-supervised data. And $\mathcal{N}$ represents the set of negative sample pairs, formed by pairing a source sample with the self-supervised data of a different sample. Therefore, for a batch of unlabeled samples with batch size $N_\text{b}$, there are $N_\text{b}$ categories in total. The contrastive features can be obtained by maximizing the similarities of positive sample pairs and minimizing the differences of negative sample pairs in the latent space. 
 	
 	To ensure the good performance of the contrastive features on downstream tasks, we integrate the multi-view learning assumption into the self-supervision, which suggests that both unlabeled sample $\mathbf{s}_\text{a}$ and self-supervised sample $\bf r$ contain sufficient information for effective task inference. We establish a Markov chain involving self-supervision and downstream tasks, expressed as $T\leftrightarrow S_\text{a} \rightarrow R$, where $S_\text{a}$ and $R$ denote the random variables of $\mathbf{s}_\text{a}$ and $\bf r$, respectively. \rev{Due to the multi-view learning assumption and the deterministic data processing of $f_\text{dc}$, we have $I(S_\text{a};T)\geq I(R;T)\geq I(S_\text{a};R;T) \geq \epsilon$, where $\epsilon$ represents the information threshold, indicating the minimum amount of task-relevant information shared between $S_\text{a}$ and $R$. Its value is evaluated by the design of $f_\text{dc}$, involving the quality of artificially constructed $R$~\cite{tsai2020self}. Specifically, if $f_\text{dc}$ significantly alters the content of $S_\text{a}$, e.g., excessive masking, $\epsilon$ decreases. When $f_\text{dc}$ applies transformations that preserve the content while only changing the style, $\epsilon$ increases. Therefore, enhancing $\epsilon$ ensures that the semantic encoder learns robust representations that retain most task-relevant information while filtering out irrelevant transformations.} 	
To this end, we apply several image augmentations involving randomness to $f_\text{dc}$, ensuring that while the unlabeled sample $\mathbf{s}_\text{a}$ is altered, its task-relevant content is preserved as much as possible. These transformations include random horizontal flip with $50\%$ probability, ColorJitter with uniformly selected adjustments, random grayscale conversion with probability $20\%$, and Gaussian blur.
	 	
	\begin{figure}[t]
		\centering
		\includegraphics[width=0.65\linewidth]{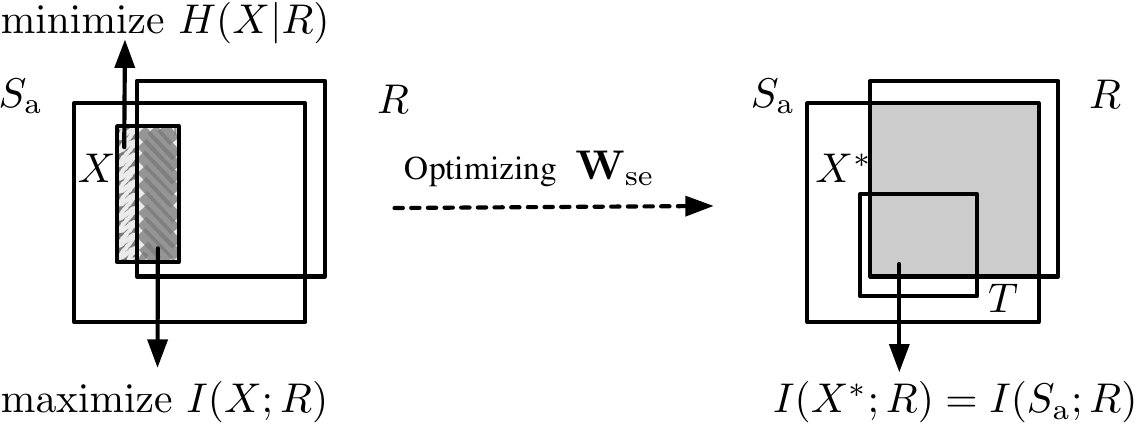}
		\caption{A Venn diagram for (P2) with self-supervision.}
		\label{ssl data venn}
	\end{figure}
	
	Building upon the self-supervision, we now reformulate problem (P1) to extract the minimal yet sufficient contrastive features from unlabeled data set $\mathcal{D}_\text{r}$ as semantic information for transmission,  yielding
	\rev{\begin{equation}\label{ssl_objective}
		\begin{split}
			(\text{P2})~\mathop {\text{maximize}}_{\mathbf{W}_\text{se}} \ & \ -\lambda H(X|R)  +  I (X;R)   \\
		  \text{subject to}   \ &\ X = f_\text{se}(S_\text{a};\mathbf{W_\text{se}}),\\
		  \ &\ R = f_\text{dc}(S_\text{a}).
		\end{split}
	\end{equation}}
	
\rev{The equivalence between (P1) and (P2) lies in their common goal of extracting task-relevant information~$T$. (P1) achieves the goal by directly guiding the recovered semantic information toward $T$ in the semantic space. Solving (P2) enables task-relevant semantic extraction is founded on the self-supervision based on the multi-view learning assumption. This assumption implies that the constructed $R$ contains most of the information that $S_\text{a}$ possesses regarding the downstream tasks. By optimizing (P2), the extracted semantic information contains the shared information between $S_\text{a}$ and $R$, which therefore inherently includes the task-relevant components.}

\rev{On the other hand, the differences between (P1) and (P2) include their approaches to learn semantic extraction and optimization variables. (P1) relies on labeled data to optimize both semantic extraction and transmission process, requiring a supervised learning approach. In contrast, (P2) employs self-supervised signals from unlabeled data to optimize the semantic extraction locally in a self-supervised manner. Notably, achieving the goal of (P1) requires an end-to-end supervised learning stage to refine the semantic representations, as described in the proposed second learning stage.}
	

	We depict a Venn diagram in Fig.~\ref{ssl data venn} to illustrate the optimization problem (P2). In the previous construction process of self-supervised data $R$ from unlabeled samples $S_\text{a}$, some data augmentation methods introduce the randomness, manifesting as the uncertainty of $R$ conditioned on the known $S_\text{a}$, i.e., the conditional entropy $H(R|S_\text{a})$.
\rev{Solving (P2) yields the optimal parameter set $\mathbf{W}^{\rm p}_{\rm se}$ and the corresponding extracted semantic information $X^\ast$.}
	When considering a downstream task $T$, these random variables form a new Markov chain, denoted by $T \leftrightarrow (S_\text{a},R)\rightarrow X^{\ast}$. 
	On one hand, the information that $S_\text{a}$ and $R$ share, known as the mutual information $I(S_\text{a};R)$, contains most content about the downstream task. Then we have $I(X^\ast;T)=I(S_\text{a};R;T)\geq \epsilon$. On the other hand, the data processing inequality implies $I(S_\text{a};T)\geq I(X^\ast;T)$. Combining both aspects, the extracted $X^\ast$ contains most task-relevant information, with information loss to $T$ denoted by $I(S_\text{a};T|R)$. 

\begin{figure*}[t]
		\centering
		\includegraphics[width=0.9\textwidth]{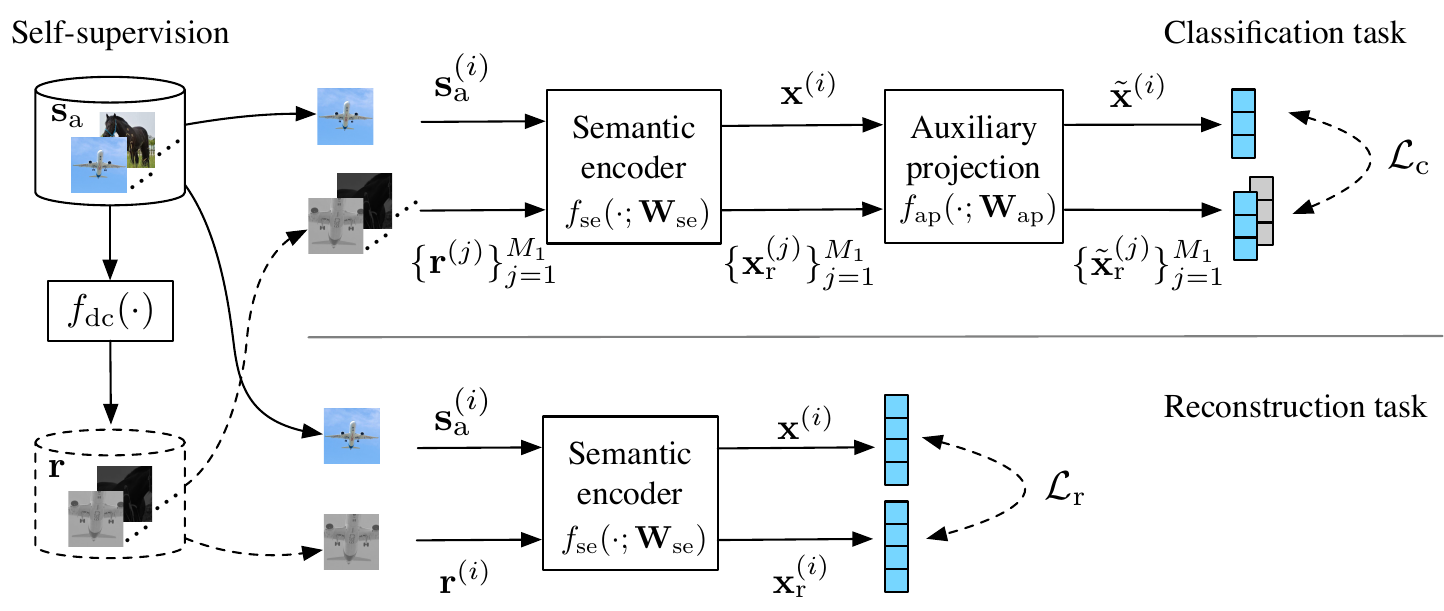}
		\caption{Pre-training for semantic encoder with pretext tasks.}
		\label{obj_pretext}
	\end{figure*}
	
\subsection{Learning Objectives for Solving (P2)}
	It is challenging to directly compute the mutual information term $I(X;R)$ and the conditional entropy term $H(X|R)$ in (P2), especially for high-dimensional data. 
	To overcome the issue, we apply an Info Noise Contrastive Estimation (InfoNCE) loss, as used in \cite{oord2018representation}, to estimate $I(X;R)$. Additionally, we introduce an approximate distribution to facilitate the computation of $H(X|R)$. These approximations render the objective function of problem (P2) tractable. We then employ two pretext tasks to formulate the approximate objective function through the learning objectives of these tasks. Specifically, a classification task aims to maximize the estimated mutual information, while a reconstruction task is responsible for minimizing the approximated conditional entropy. The optimization of both learning objectives is carried out by using on-device networks with the background of self-supervision for different tasks as follows. 

	\subsubsection{Classification Task for Maximizing $I(X;R)$}
	We use InfoNCE to formulate a learning objective in the classification task, aiming to maximize $I(X;R)$. InfoNCE serves to enhance the dependency and contrast between the learned representations and self-supervised data. It has been empirically shown to provide a lower bound on the mutual information between the learned representations and the self-supervised data\cite{oord2018representation}. In InfoNCE, we apply a similarity metric to quantify the agreement between a pair of examples within the latent space, and use a softmax function to convert the similarity scores into a probability distribution, where higher probabilities are assigned to positive sample pairs and lower probabilities to negative ones.
	
	We define the classification task using InfoNCE for the on-device network, as depicted in Fig. \ref{obj_pretext}. In this classification task, given the unlabeled sample $\mathbf{s}^{(i)}_\text{a}$ and a set of $M_1$ self-supervised data $\{\mathbf{r}^{(j)}\}^{M_1}_{j=1}$ as obtained in the last subsection, the goal is to identify $\mathbf{r}^{(i)}$ in the set that matches with $\mathbf{s}^{(i)}_\text{a}$. Within self-supervision, correct classification results can be readily obtained by examining the set of positive sample pairs. 
	
	To enhance the classification score, it is crucial to maximize the joint probability of each positive sample pair. This is achieved by assigning higher probabilities to positive sample pairs in the InfoNCE. We leverage the on-device network to obtain low-dimensional vectors, which are then used for measuring the similarity. Specifically, the semantic encoder network $f_\text{se}(\cdot;\mathbf{W}_\text{se})$ is applied to $\mathbf{s}^{(i)}_\text{a}$ and $\{\mathbf{r}^{(j)}\}^{M_1}_{j=1}$, extracting respective semantic information $\mathbf{x}^{(i)}$ and $\{\mathbf{x}^{(j)}_\text{r}\}^{M_1}_{j=1}$. The subsequent auxiliary projection $f_\text{ap}(\cdot;\mathbf{W}_\text{ap})$ generates lower-dimensional vectors $\tilde{\mathbf{x}}^{(i)}$ and $\{\tilde{\mathbf{x}}^{(j)}_\text{r}\}^{M_1}_{j=1}$. For a batch of data samples $\{(\mathbf{s}^{(i)}_\text{a})^{N_\text{b}}_{i=1}\}$ with batch size $N_\text{b}$ from the remote dataset $\mathcal{D}_\text{r}$, we employ the InfoNCE loss with similarity measured by inner product. This loss forms the basis of the learning objectives for updating $f_\text{se}(\cdot;\mathbf{W}_\text{se})$ and $f_\text{ap}(\cdot;\mathbf{W}_\text{ap})$, which is written as
	\vspace{-6pt}
	\begin{align}\label{eq:pre_cl}
		\mathop{\text{maximize}}_{\mathbf{W}_\text{se},\mathbf{W}_\text{ap}}~\mathcal{L}_\text{c} \triangleq \frac{1}{N_\text{b}} \sum_{i=1}^{N_\text{b}} \log \frac{e^{ \tilde{\bf x}^{(i)}\cdot \tilde{\bf x}^{(i)}_\text{r}} }{\sum_{j=1}^{N_\text{b}} e^{\tilde{\bf x}^{(i)}\cdot \tilde{\bf x}^{(j)}_\text{r}}} .	
	\end{align}
	By optimizing the objective in \eqref{eq:pre_cl}, we learn $f_\text{se}$ for extracting the shared information between ${\bf s}_\text{a}$ and $\bf r$, containing most task-relevant information.
	\begin{algorithm}[t]
		\SetAlgoLined
		\DontPrintSemicolon
		\KwIn{$\mathcal{D}_\text{r}$, $\lambda$, $N_\text{b}$, and learning rate $\eta$.}
		\For{$t=1:\lfloor M_1/N_{\rm b}\rfloor$}{
	        \For{sampled minibatch $\{\mathbf{s}^{(k)}_\text{\rm a}\}^{N_\text{\rm b}}_{k=1}$}{
				\For{$k=1:N_\text{\rm b}$}{
					Generate self-supervised data: $\mathbf{r}^{(k)}=f_\text{dc}(\mathbf{s}^{(k)}_\text{a})$. \\
					Extract semantic information: $\mathbf{x}^{(k)}=f_\text{se}(\mathbf{s}^{(k)}_\text{a};\mathbf{W}_\text{se})$; $\mathbf{x}^{(k)}_\text{r}=f_\text{se}(\mathbf{r}^{(k)};\mathbf{W}_\text{se})$.\\
					Project semantic information:
					$\mathbf{\tilde{x}}^{(k)}=f_\text{ap}(\mathbf{x}^{(k)};\mathbf{W}_\text{ap})$; $\mathbf{\tilde{x}}^{(k)}_\text{r}=f_\text{ap}(\mathbf{x}^{(k)}_\text{r};\mathbf{W}_\text{ap})$.\\
				}
				Parameter updating by minimizing $\mathcal{L}_\text{pre}$:\\
				$\mathbf{W}^{(t+1)}_{\rm se} \leftarrow \mathbf{W}^{(t)}_{\rm se}-\eta \bigtriangledown_{\mathbf{W}_{\rm se}} \mathcal{L}_{\rm pre}$,\\
	$\mathbf{W}^{(t+1)}_{\rm ap} \leftarrow \mathbf{W}^{(t)}_{\rm ap}-\eta \bigtriangledown_{\mathbf{W}_{\rm ap}} \mathcal{L}_{\rm c}$,
				}
				}
			\KwOut{$\mathbf{W}^\text{p}_\text{se}$.}
		\caption{Self-supervised Pre-training}
	\label{SSLTA_1}
	\end{algorithm}
	
	\subsubsection{Reconstruction Task for Minimizing $H(X|R)$}
	We define the reconstruction task inspired by the computation of conditional entropy, that is $H(X|R)=-\mathbb{E}_{p(\mathbf{x},\mathbf{r})}[\log p(\mathbf{x}|\mathbf{r})]$. The task aims to prompt the self-supervised data to reconstruct the semantic information, via maximizing $\mathbb{E}_{p(\mathbf{x},\mathbf{r})}[\log p(\mathbf{x}|\mathbf{r})]$. To avoid intractability in computing the log conditional likelihood, we introduce a distribution $q(\mathbf{x}|\mathbf{r})$ as an alternative approximate to the true distribution $p(\mathbf{x}|\mathbf{r})$, where due to the Kullback-Leibler divergence, maximizing the log conditional likelihood of $q(\mathbf{x}|\mathbf{r})$ serves as a lower bound for $\mathbb{E}_{p(\mathbf{x},\mathbf{r})}[\log p(\mathbf{x}|\mathbf{r})]$\cite{xie2023disentangled}. 
	
	The choice of the distribution form for $q(\mathbf{x}|\mathbf{r})$ dictates the formulation of the learning objective. When selecting the distribution $q(\mathbf{x}|\mathbf{r})$, we account for the semantic difference between $\mathbf{s}_\text{a}$ and $\bf r$ rather than their distinctions in the lower dimensional space. This is because the auxiliary projection inevitably introduces information loss, contradicting the reconstruction goal. As depicted in the lower part of Fig.~\ref{obj_pretext}, for any positive sample pair $(\mathbf{s}^{(i)}_\text{a},\mathbf{r}^{(i)})$, we aim to guarantee the consistency between $\mathbf{x}^{(i)}$ and $\mathbf{x}^{(i)}_\text{r}$, aligning with the idea of our self-supervision. Motivated by the Gaussian approximation applied in image reconstruction\cite{choi2019neural}, we choose $q(\mathbf{x}|\mathbf{r}) = \mathcal{CN}(\mathbf{x}|f_\text{se}(\mathbf{r};\mathbf{W}_\text{se}),\mathbf{I})$, which implies that any ${\bf x}^{(i)}$ of the training data ${\bf s}_\text{a}^{(i)}$ follows the Gaussian distribution with mean given by ${\bf x}^{(i)}_\text{r}=f_\text{se}(\mathbf{r}^{(i)};\mathbf{W}_\text{se})$ and variance of an identity matrix $\bf I$ with dimension $L_\text{s}\times L_\text{s}$.
	Then the learning objective for the reconstruction task is formulated by maximizing the log likelihood of $q(\mathbf{x}|\mathbf{r})$,
	\vspace{-3mm}
	\begin{align}\label{eq:pre_ip}
		\mathop{\text{maximize}}_{\mathbf{W}_\text{se}}~\mathcal{L}_\text{r} &\triangleq \ \frac{1}{N_\text{b}} \sum_{i=1}^{N_\text{b}} \ln \mathcal{CN}(\mathbf{x}^{(i)}|\mathbf{x}^{(i)}_\text{r},\mathbf{I}) \nonumber \\
		&\equiv -\frac{1}{N_\text{b}} \sum_{i=1}^{N_\text{b}} \| \mathbf{x}^{(i)}-\mathbf{x}^{(i)}_\text{r} \|^2_2,
	\end{align}
	where the last equivalence is due to omitting the constants that do not affect the optimization. 
	The semantic encoder $f_\text{se}$ is trained to encourage $\bf x$ to discard information pertaining to $\bf r$, which primarily refers to the information introduced by the self-supervised data construction strategies and is thus irrelevant to the task.

	Finally, we unify the two learning objectives in the local self-supervised learning stage as $\mathcal{L}_\text{pre}=- (\lambda \mathcal{L}_\text{r} + \mathcal{L}_\text{c})$. By minimizing $\mathcal{L}_\text{pre}$, we learn $f_\text{se}$ to extracts most task-relevant information. The training algorithm is summarized in Algorithm \ref{SSLTA_1}. The auxiliary projection $f_\text{ap}(\cdot;\mathbf{W}_\text{ap})$ is to generate lower-dimensional vectors for similarity comparison. Once the training is accomplished, this projection will be discarded.

	\subsection{Integrating $f_\text{\rm se}$ in SLSCom}
	The proposed SLSCom is designed with a task-oriented principle, emphasizing the significant contribution of semantic information to downstream tasks. The semantic encoder $f_\text{se}(\cdot;\mathbf{W}^{\text{p}}_\text{se})$ has been self-supervisedly learned, enabling it to extract contrastive features containing most task-relevant information. This end-to-end supervised learning stage is to leverage the extracted semantic information for satisfactory task inference performance. We employ two typical methods to integrate the pre-trained $f_{\rm se}$ within SLSCom. 
	\subsubsection{End-to-end Fine-tuning} $f_{\rm se}$, JSCC network, and semantic decoder network are jointly trained for end-to-end inference accuracy. This method leverages the adaptability of the pre-trained model by fine-tuning it to suit the downstream tasks. 
	\subsubsection{Frozen $f_{\rm se}$ Integration} We explore freezing the parameters of $f_{\rm se}$ during the end-to-end learning process. This method avoids retraining of $f_{\rm se}$ and allows for a direct evaluation of the quality of the pre-trained $f_{\rm se}$.
	
	In order for robust inference performance under wireless fading channels, we introduce the MSE loss function and cross-entropy (CE) loss function for semantic transmission and interpretation, respectively. Specifically, the MSE loss function applies to the JSCC network, which quantifies the semantic distortion between the semantic information $\bf x$ and the recovered semantic information $\bf \hat{x}$. For a batch of data samples $\{(\mathbf{s}^{(i)}_\text{t},\mathbf{t}^{(i)})^{N_\text{b}}_{i=1}\}$ with batch size $N_\text{b}$ from the labeled dataset $\mathcal{D}_\text{s}$, the MSE loss function is express by
	\vspace{-6pt} 
	\begin{align}\label{eq:app_mse}
		\mathcal{L}_\text{mse}\triangleq  \ \frac{1}{N_\text{b}} \sum_{i=1}^{N_\text{b}} \|\mathbf{x}^{(i)}-\hat{\mathbf{x}}^{(i)} \|^2_2.
	\end{align}
	Moreover, the semantic decoder network is tasked with interpreting the semantic information for downstream tasks. Take the image classification task as an example, the semantic decoder network is trained using the CE loss function, which measures the similarity in the probabilities between the predicted result and the ground truth. It follows
	\vspace{-2mm}
	\begin{align}\label{eq:app_ce}
		\mathcal{L}_\text{ce}  \triangleq \ -\frac{1}{N_\text{b}}\sum_{i=1}^{N_\text{b}} \mathbf{t}^{(i)}\cdot \log \mathbf{y}^{(i)},
	\end{align}
	where $\mathbf{t}$ and $\mathbf{y}$ are column vectors of equal length $N_\text{cls}$, with each element representing the probability of belonging to a specific class.

	\begin{algorithm}[t]
		\SetAlgoLined
		\DontPrintSemicolon
		\KwIn{$\mathcal{D}_\text{s}$, $\mu$, $N_\text{b}$, and learning rate $\eta$; Learned $f_\text{se}(\cdot;\mathbf{W}_\text{se}^{\text{p}})$; Reference symbol $\mathbf{Z}_\text{p}$.} 
		\For{$t=1:\lfloor M_2/N_{\rm b}\rfloor$}{
	        \For{sampled minibatch $\{(\mathbf{s}^{(k)}_\text{\rm t},\mathbf{t}^{(k)})^{N_\text{\rm b}}_{k=1}\}$}{
				\For{$k=1:N_\text{\rm b}$}{
					Extract semantic information: $\mathbf{x}^{(k)}=f_\text{\rm se}(\mathbf{s}^{(k)}_\text{t};\mathbf{W}^{(t)}_\text{se})$.\\
					Receive $\mathbf{\hat{x}}^{(k)}$ by $\eqref{eq:encoded}$ and $\eqref{eq:decoded}$ with OFDM transceiver and $\mathbf{Z}_\text{p}$.\\
					Infer $\mathbf{y}^{(k)}$ by $\eqref{eq:inference}$. \\		
				}
				Parameter updating by minimizing $\mathcal{L}_\text{app}$:\\
				$\mathbf{W}^{(t+1)}_{\rm se} \leftarrow \mathbf{W}^{(t)}_{\rm se}-\eta \bigtriangledown_{\mathbf{W}_{\rm se}} \mathcal{L}_{\rm app}$,\\
	$\mathbf{W}^{(t+1)}_{\rm je} \leftarrow \mathbf{W}^{(t)}_{\rm je}-\eta \bigtriangledown_{\mathbf{W}_{\rm je}} \mathcal{L}_{\rm app}$,\\
	$\mathbf{W}^{(t+1)}_{\rm jd} \leftarrow \mathbf{W}^{(t)}_{\rm jd}-\eta \bigtriangledown_{\mathbf{W}_{\rm jd}} \mathcal{L}_{\rm app}$,\\
	$\mathbf{W}^{(t+1)}_{\rm sd} \leftarrow \mathbf{W}^{(t)}_{\rm sd}-\mu\eta \bigtriangledown_{\mathbf{W}_{\rm sd}} \mathcal{L}_{\rm ce}$.
			}
			}
			\KwOut{ $\mathbf{W}^\text{f}_\text{se}$, $\mathbf{W}^\text{f}_\text{je}$, $\mathbf{W}^\text{f}_\text{jd}$, and $\mathbf{W}^\text{f}_\text{sd}$.}
		\caption{Supervised Fine-tuning}
		\label{SSLTA_2}
	\end{algorithm}
	
	The overall learning objective is to minimize the weighted sum of these two loss function, i.e.,
	\begin{align}\label{eq:app}
		\mathop{\text{minimize}}_{\mathbf{W}_\text{se}^\text{p},\mathbf{W}_\text{je},\mathbf{W}_\text{jd},\mathbf{W}_\text{sd}}~\mathcal{L}_\text{app} = \mathcal{L}_\text{mse} + \mu \mathcal{L}_\text{ce},
	\end{align}
	where $\mu >0$ is a hyper-parameter. As training converges, the optimized parameter sets, including $\mathbf{W}^{\rm f}_{\rm se}$, $\mathbf{W}^{\rm f}_{\rm je}$, $\mathbf{W}^{\rm f}_{\rm jd}$, and $\mathbf{W}^{\rm f}_{\rm sd}$, are obtained. The learned semantic encoder and decoder focus on task-relevant semantic extraction and interpretation, while the learned JSCC network compress and decompress semantic information, mitigating the effects of channel distortion for accurate task inference. 
	 
	The training procedure is outlined in Algorithm~\ref{SSLTA_2}. It is important to note that the initial parameter set of the semantic encoder, $\mathbf{W}^{(1)}_{\rm se}$, is initialized to $\mathbf{W}^{\rm p}_{\rm se}$, which represents the pre-trained parameter set from the local self-supervised learning stage. Furthermore, when we choose to freeze the parameters of $f_{\rm se}$, we have $\mathbf{W}^{\rm f}_{\rm se} = \mathbf{W}^{(t)}_{\rm se}=\mathbf{W}^{\rm p}_{\rm se}$ for any $t$.  
	

\section{Simulation Results}
	
	In this section, we verify the effectiveness of the proposed framework on image classification tasks. We compare the performance of the proposed framework with existing DL methods and typical digital coding approaches across varying labeled dataset sizes and SNR conditions. Additionally, we investigate performance under different numbers of unlabeled samples and the impact of the distribution of these unlabeled samples. Finally, through ablation studies, we discuss the impacts of the reconstruction pretext task and color-related transformations, as well as the necessity of the auxiliary projection layer.	
	

	\begin{table*}[ht]
	\caption{DL Baselines and the Proposed SLSCom}
	\centering
	\label{tab:baselines}
	\renewcommand\arraystretch{1.5}
	\resizebox{\linewidth}{!}{%
	\begin{threeparttable}
	\begin{tabular}{l|l|cccccc}
	\toprule
	\multirow{3}{*}{Baselines} & \multicolumn{1}{c|}{\multirow{3}{*}{Training methods}} & \multicolumn{4}{c}{\# Parameters/FLOPs}                                                                                                                                                                                                                                         \\ \cline{3-6} 
	                           & \multicolumn{1}{c|}{}                                  & \multicolumn{2}{c|}{$(\text{CR}_\text{s}, \text{CR}_\text{c})=(\frac{2}{3},\frac{1}{4})$}                                                                    & \multicolumn{2}{c}{$(\text{CR}_\text{s}, \text{CR}_\text{c})=(\frac{1}{6},\frac{1}{8})$}                                                 \\ \cline{3-6} 
	                           & \multicolumn{1}{c|}{}                                  & \multicolumn{1}{c|}{Pre-training}              & \multicolumn{1}{c|}{Fine-tuning}            & \multicolumn{1}{c|}{Pre-training}              & Fine-tuning             \\ \hline 
	RSCom                       & Training from scratch                                  & \multicolumn{1}{c|}{\multirow{2}{*}{$-$}}                         & \multicolumn{1}{c|}{\multirow{5}{*}{$29.06$ M/$1.82$ G}}                 & \multicolumn{1}{c|}{\multirow{2}{*}{$-$}}                          & {\multirow{5}{*}{$21.63$ M/$1.19$ G}}                 \\ \cline{1-2}
	TSCom                        & Training with pre-trained ResNet model                  & \multicolumn{1}{c|}{}  & \multicolumn{1}{c|}{} & \multicolumn{1}{c|}{}  &  \\ \cline{1-3} \cline{5-5} 
	SLSCom\tnote{$\dag$}                      & The proposed framework & \multicolumn{1}{c|}{\multirow{2}{*}{$27.95$ M/$1.32$ G}} & \multicolumn{1}{c|}{}                                               & \multicolumn{1}{c|}{\multirow{2}{*}{$21.60$ M/$1.16$ G}} &                         \\ \cline{1-2}
	SLSCom\tnote{$\dag$}~w/o R                     & SLSCom without reconstruction pretext task          & \multicolumn{1}{c|}{}                          & \multicolumn{1}{c|}{}                         & \multicolumn{1}{c|}{}                          &                         \\  \cline{1-3} \cline{5-5}
	SLSCom\tnote{$\dag$} w/o AP                        & SLSCom without auxiliary projection                 & \multicolumn{1}{c|}{$23.51$ M/$1.31$ G}  & \multicolumn{1}{c|}{} & \multicolumn{1}{c|}{$21.28$ M/$1.16$ G}  &  \\ \bottomrule
	\end{tabular}%
	\begin{tablenotes}
		\footnotesize
		\item[$\dag$] Proposed method in this paper.
	\end{tablenotes}
	\end{threeparttable}
	}
	\vspace{-0.3cm}
	\end{table*}
	
	\subsection{Simulation Setup}
	\subsubsection{Channel and OFDM Parameters}
	We conduct all simulations under multipath wireless channels. Channel parameters are set as: $L_\text{h}=8$ and each path follows $h_l \sim \mathcal{CN}(0, p_l e^{-\frac{1}{4}})$ for $l=1,2,\dots, L_\text{h}$, where $p_l$ is a weight that ensures the sum of weighted $p_1,\ldots,p_{L_\text{h}}$ equals $1$. The OFDM parameters include $K=64$, $N_\text{s}=6$, $N_\text{p}=2$, and $L_\text{cp}=16$. \rev{For the pilot sequence, it is a deterministic sequence in our simulations, which is randomly generated as a binary sequence. Once generated, the pilot sequence remains unchanged throughout the simulations.}
	\subsubsection{Datasets}
	We use the CIFAR10 dataset\cite{krizhevsky2009learning}, and Street View House Numbers (SVHN) dataset\cite{netzer2011reading}, and Flowers recognition dataset\cite{mamaev2021flowers} for the image classification task. CIFAR10 consists of $60,000$ images with a pixel size of $32\times 32$. SVHN contains a training set of $73,257$ images and a testing set of $26,032$ images, also with a pixel size of $32\times 32$. Flowers recognition includes $4,242$ flower images, with resolutions approximately $320\times 240$ pixels. \rev{The total number of available training samples, $M$, is $50,000$ for CIFAR10 and $73,257$ for SVHN. The number of unlabeled samples, $M_1$, is set to $32,000$ for both CIFAR10 and SVHN. For Flower recognition dataset, both $M_1$ and $M$ are set to $2,814$, accounting for approximately $66\%$ of the total dataset.}
	
	\subsubsection{NN Implementation}	
	We evaluate two types of NN structures for all methods, represented by different CR pairs, as shown in Table~\ref{tab:baselines}. For the CR pair $(\frac{2}{3},\frac{1}{4})$, we use ResNet-$50$ as the semantic encoder, with the JSCC network configured by setting the initial filter count $N_\text{f}$ to $64$, the input filter count $N_\text{fc}$ to $12$, and the sampling factor $N_\text{sa}$ to $2$. For the CR pair $(\frac{1}{6},\frac{1}{8})$, ResNet-$34$ is used, and the parameters are adjusted to $N_\text{f} = 32$, $N_\text{fc} = 6$, and $N_\text{sa} = 1$. Both ResNet-$50$ and ResNet-$34$ are modified by removing their last FC layers. Unless otherwise specified, the NN structure with the CR pair $(\frac{2}{3},\frac{1}{4})$ is used.
	
	All approaches are trained using the Adam optimizer with a learning rate of $10^{-4}$ ($\beta_1=0.5, \beta_2=0.99$). The batch size is set to $128$. \rev{The training epochs are set to $200$ for local self-supervised learning and $60$ for end-to-end supervised learning.} To better evaluate the performance, we train each approach $10$ times and test each well-trained model $10$ times, followed by calculating the average performance.
	
	\rev{We determine the hyperparameters in the learning objectives by evaluating the validation performance on both pretext and downstream tasks. Specifically, we assess the learned semantic encoder $f_\text{se}$ using various $\lambda$, specifically $\lambda \in \{0.05, 0.1, 0.15, 0.2\}$. The experiments focus on identifying matched self-supervised data from a batch, with performance measured using Top-1 accuracy for the pretext classification task. The optimal $\lambda$ is selected based on the best performance on this validation set. Additionally, $\mu$ is determined based on the validation performance of the downstream classification tasks.} In particular, $\lambda = 0.15$ when $\text{CR}_\text{s} = \frac{2}{3}$, $\lambda = 0.1$ when $\text{CR}_\text{s} = \frac{1}{6}$, and $\mu=1$.
	
	\subsubsection{Baselines and the Proposed SLSCom} We consider several typical DL methods and digital coding schemes, with the DL baselines are summarized in Table~\ref{tab:baselines}.
	\begin{enumerate}[$\bullet$]
		\item RSCom: We perform end-to-end learning using labeled samples, where NNs are trained from scratch with all learnable parameters initialized at \textit{random}.
		\item TSCom and its frozen $\mathbf{W}^\text{p}_\text{se}$ variant: We load a pre-trained ResNet model as the semantic encoder and \textit{transfer} it for end-to-end learning, following the approach in \cite{lee2022seq2seq}. The variant, where the weights of the pre-trained model $\mathbf{W}^\text{p}_\text{se}$ are kept fixed during training, is referred to as TSCom with frozen weights (TSCom w FW). 
		\item SLSCom, its ablation of reconstruction pretext task (SLSCom w/o R) and its ablation of auxiliary projection (SLSCom w/o AP): The proposed framework  uses $32,000$ unlabeled samples for self-supervised pre-training if not specified otherwise. Additionally, we examine its variants of frozen $\mathbf{W}^\text{p}_\text{se}$ (SLSCom w FW) and data distribution shifts (SLSCom w DS). \rev{Within the SLSCom w DS scheme, we first pre-train the semantic encoder using $32,000$ training samples from the SVHN dataset. Subsequently, training samples from the CIFAR10 dataset are employed to jointly train the semantic encoder alongside the other modules.}
		\item Conventional method with digital coding: We use joint photographic experts group (JPEG) for image source coding, low-density parity-check code (LDPC) for channel coding, $8$-quadrature amplitude modulation (QAM), OFDM technology, least squares (LS) channel estimation, and minimum mean square error (MMSE) equalization. A classifier is trained using $40,000$ labeled samples, with the NN structure being a concatenation of the semantic encoder and decoder networks as shown in Fig.~\ref{slscArchitecture}.
		\item Hybrid semantic and digital coding: We use $8$-bit $\mu$-law quantization, LDPC coding, $8$-QAM modulation, and OFDM to transmit the semantic information. At the receiver, the LS channel estimation and MMSE equalization are used to recover the semantic information. The semantic encoder and decoder networks are trained following the proposed learning approach, though without involving the DJSCC network. Specifically, $32,000$ unlabeled samples are used for local self-supervised learning, followed by joint training of the semantic encoder and decoder networks with $40,000$ labeled samples.
	\end{enumerate}
 	We use an LDPC code with a coding rate of $2/3$ and a parity-check matrix of size $864\times 2304$. The semantic encoder and decoder network structures follow the configuration indicated by CR pair $(\frac{2}{3},\frac{1}{4})$.

	\begin{figure*}[t]
		\centering
		\hspace{-0.3cm}
		\subfigure[CR pair ($\frac{2}{3}$, $\frac{1}{4}$)]{
		\label{r50_on_cifar}
        \includegraphics[width=0.42\linewidth]{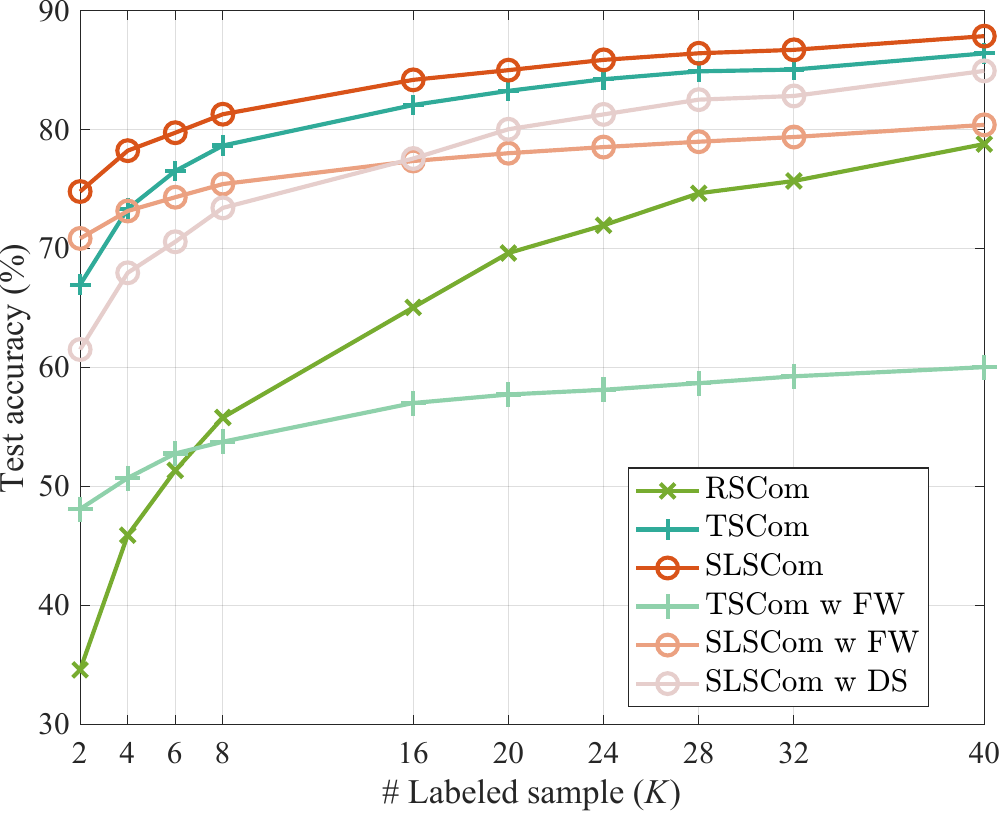}
		}
		\subfigure[CR pair ($\frac{1}{6}$, $\frac{1}{8}$)]{
		\label{r34c_on_cifar}
		\includegraphics[width=0.42\linewidth]{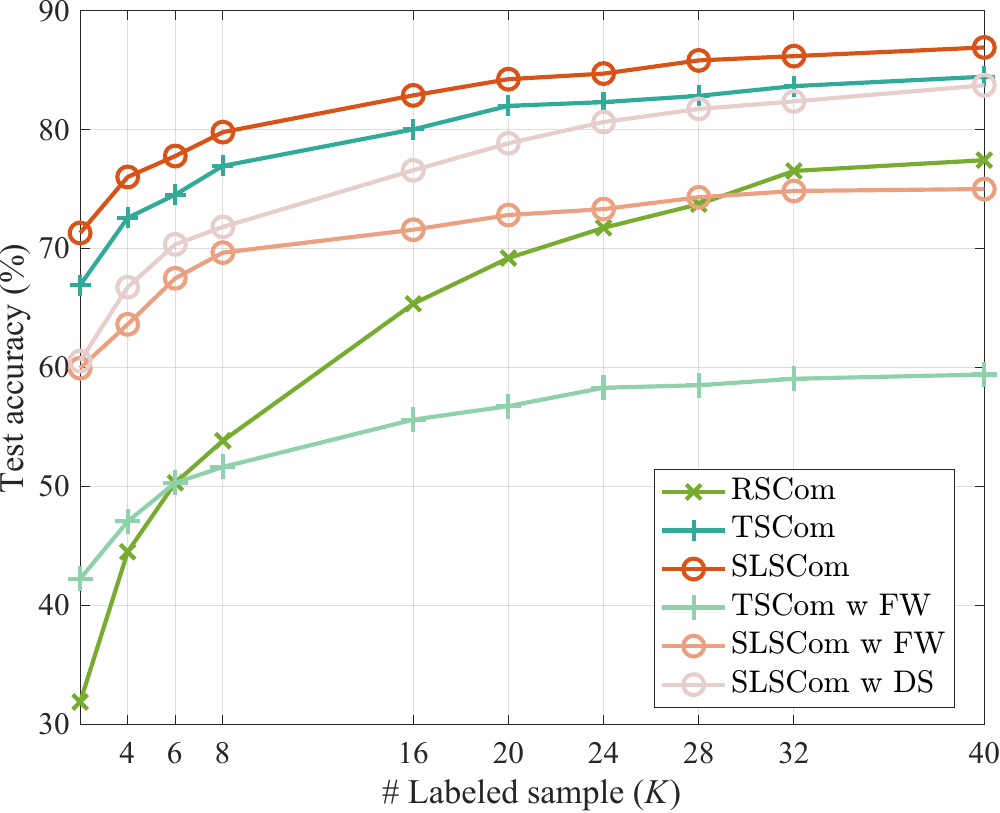}
		}
		\caption{Comparison of test accuracy on CIFAR10 versus the number of labeled samples.}
		\label{cifar10_acc}
	\vspace{-0.2cm}
	\end{figure*}

	\subsection{Performance under Varying Labeled Data Set Sizes}
	To validate the superiority of the proposed framework, we compare its classification task inference performance with existing DL methods across varying labeled samples.
	The classification performance is commonly measured using the Top-1 accuracy metric, which quantifies the percentage of samples in which the predicted class with the highest probability matches the ground truth. 
	
	Overall, the proposed framework achieves superior inference performance, particularly with a small set of labeled data. The test accuracy curves for the CIFAR10 classification task are shown in Fig.~\ref{cifar10_acc}, where all methods trained and tested at $0$~dB, unless otherwise specified. It is observed that SLSCom consistently outperforms other methods across varying NN structures and labeled data set sizes. The achieved performance improvements demonstrate the strength of the proposed self-supervised learning method, as the pre-trained model enhances end-to-end inference accuracy by extracting the semantic information that is relevant to the downstream task and robust against channel noise. Its effectiveness is further highlighted by evaluating the training strategies of the frozen $\mathbf{W}^{\rm p}_\text{se}$ integration and data distribution shifts. 
	While TSCom benefits from being trained on the large and diverse ImageNet dataset, this pre-training introduces a data distribution shift because the ImageNet dataset is not specifically tailored to the downstream task we evaluated. In contrast, SLSCom is pre-trained using data directly related to the downstream task, albeit unlabeled, and with only approximately $1/40$ of the data volume of ImageNet. The distribution shift can lead to a mismatch in feature extraction and therefore cause TSCom to achieve inferior performance, which is more severe in the frozen $\mathbf{W}^{\rm p}_\text{se}$ strategy, i.e., TSCom~w~FW. However, TSCom still outperforms SLSCom's variant of data distribution shifts, i.e., SLSCom w DS, where we use the SVHN dataset for local self-supervised learning. This is likely because TSCom's extensive exposure to the large-scale ImageNet provides a generalizable foundation that compensates for the distribution shift to some extent. Furthermore, the classification accuracy of RSCom, with weights initialized arbitrarily, improves significantly as the number of labeled samples increases. This sensitivity to data volume underscores the significant dependence on labeled samples when training from scratch, a common practice in most semantic communication studies. When shifting from a deeper NN structure to a lighter one, all methods experience a certain degree of performance degradation due to the reduction in layers and filters. \rev{This degradation is more pronounced in methods with frozen weights, as the semantic extraction module cannot adapt to compensate for the lighter NN's reduced representational capacity. TSCom~w~FW, benefiting from the large-scale pre-training, has greater adaptability during this shift. SLSCom~w~FW has limited ability to recover the loss in feature complexity when using the lighter network.}

	\begin{figure}[t]
		\centering
		\includegraphics[width=0.6\linewidth]{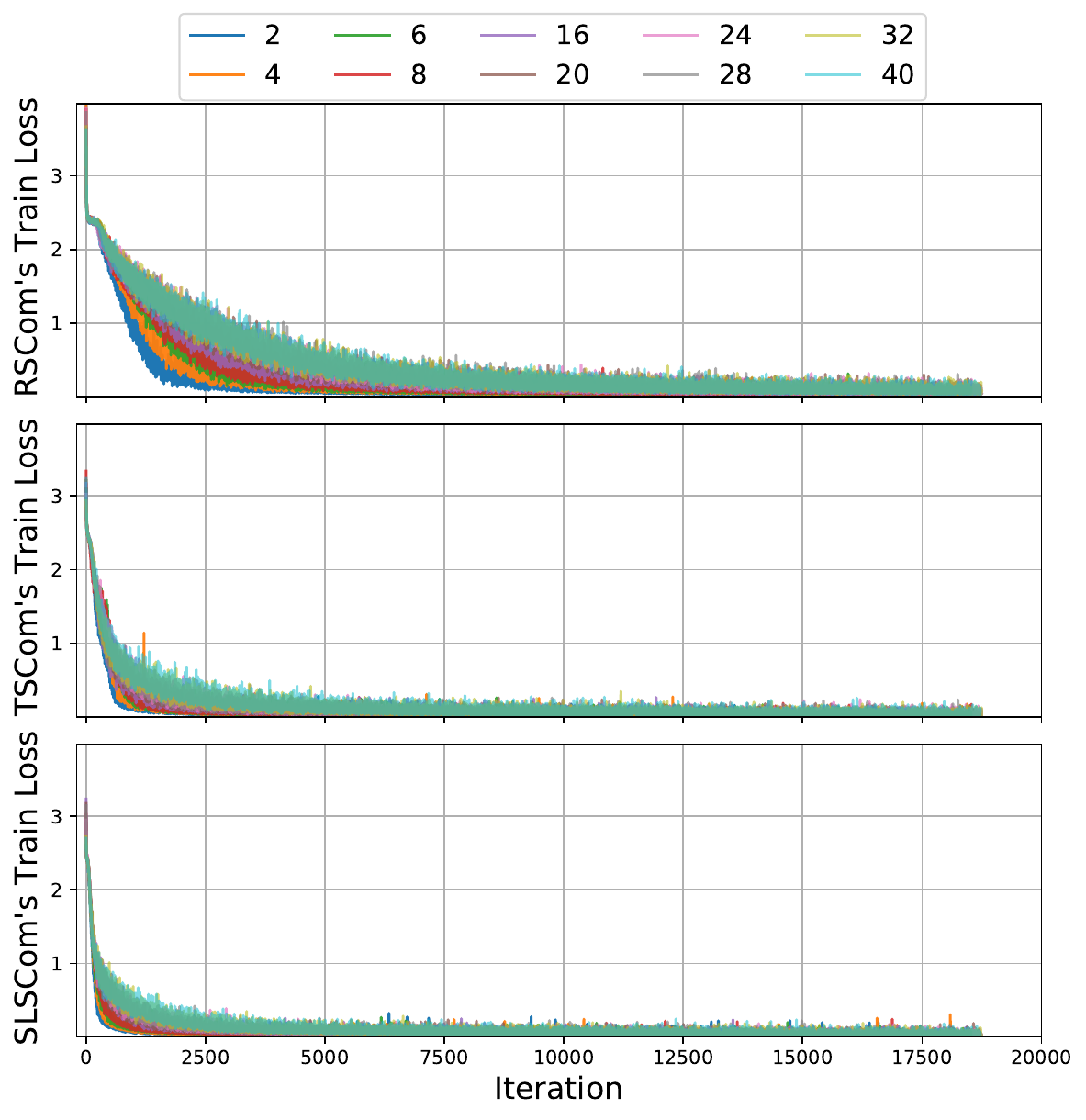}
		\caption{Training loss on CIFAR10 for RSCom, TSCom, and SLSCom with varying numbers of labeled samples.}
		\label{train_loss}
	\end{figure}	
\rev{To elaborate on the convergence behavior for training, we present in Fig.~\ref{train_loss} the training loss of RSCom, TSCom, and SLSCom for different $M_2$ under an identical number of parameter update iterations. The convergence behavior is significantly influenced by $M_2$. For small values of $M_2$, e.g., $2,000$ and $4,000$, the training loss exhibits a steep initial decline, indicating rapid adaptation. However, this early-stage performance is often misleading, as the model tends to overfit due to the scarcity of labeled data. Consequently, the final converged loss remains relatively high, as the limited labeled data fails to provide sufficient supervision, restricting the model's ability to generalize effectively to unseen data. As $M_2$ increases, the training loss reduces gradually before converging to a much better (lower) value, reflecting improved generalization benefiting from richer supervision with more labeled data. Among the three methods, RSCom starts with a higher initial loss, requiring more iterations to achieve a comparable loss level. This is attributed to RSCom's lack of the prior knowledge that both TSCom and SLSCom have already accumulated. In contrast, TSCom and SLSCom converge more efficiently, achieving better convergence loss. Notably, SLSCom demonstrates faster convergence, particularly under small $M_2$, showcasing the advantages of the proposed local self-supervised learning strategy. We wish to emphasize that the number of epochs is set to $60$ for consistent and efficient training across all schemes, resulting in varying numbers of parameter update iterations for different $M_2$.}

	
	\begin{table*}[t]
		\centering
		\caption{Test Accuracy Comparison on CIFAR10 versus the Number of Labeled Samples with Training SNR $-4$ dB}
		\label{tab:acc_nlabel_4dB}
		\resizebox{\textwidth}{!}{%
		\begin{threeparttable}
		\begin{tabular}{c|c|c|c|c|c|c|c}
		\toprule
		\# Labeled sample $(K)$ & $2$ & $4$ & $8$ & $16$ & $24$ & $28$ & $40$ \\ \hline
		RSCom & $23.41~(32.28\%)$ & $34.29~(25.29\%)$ & $52.14~(6.54\%)$ & $64.31~(1.14\%)$  & $70.62~(1.85\%)$ & $73.87~(1.04\%)$ & $79.02~(-0.29\%)$ \\ \hline
		TSCom & $40.55~(39.44\%)$ & $65.95~(10.08\%)$ & $77.81~(1.08\%)$ & $81.74~(0.40\%)$  & $83.66~(0.69\%)$ & $83.71~(1.40\%)$ & $85.88~(0.61\%)$ \\ \hline
		SLSCom & $60.66~(18.90\%)$ & $74.72~(4.49\%)$ & $80.17~(1.39\%)$ & $83.18~(1.20\%)$ & $85.06~(0.93\%)$ & $85.73~(0.80\%)$ & $87.30~(0.64\%)$ \\ \bottomrule
		\end{tabular}%
		\begin{tablenotes}
			\footnotesize
			\item[$\ast$] A positive/negative percentage reflects a reduction/improvement in accuracy under training SNR $-4$~dB compared to training SNR $0$~dB.
		\end{tablenotes}
		\end{threeparttable}
		}
	\end{table*}

	\begin{figure}[t]
    \centering
    \includegraphics[width=0.75\linewidth]{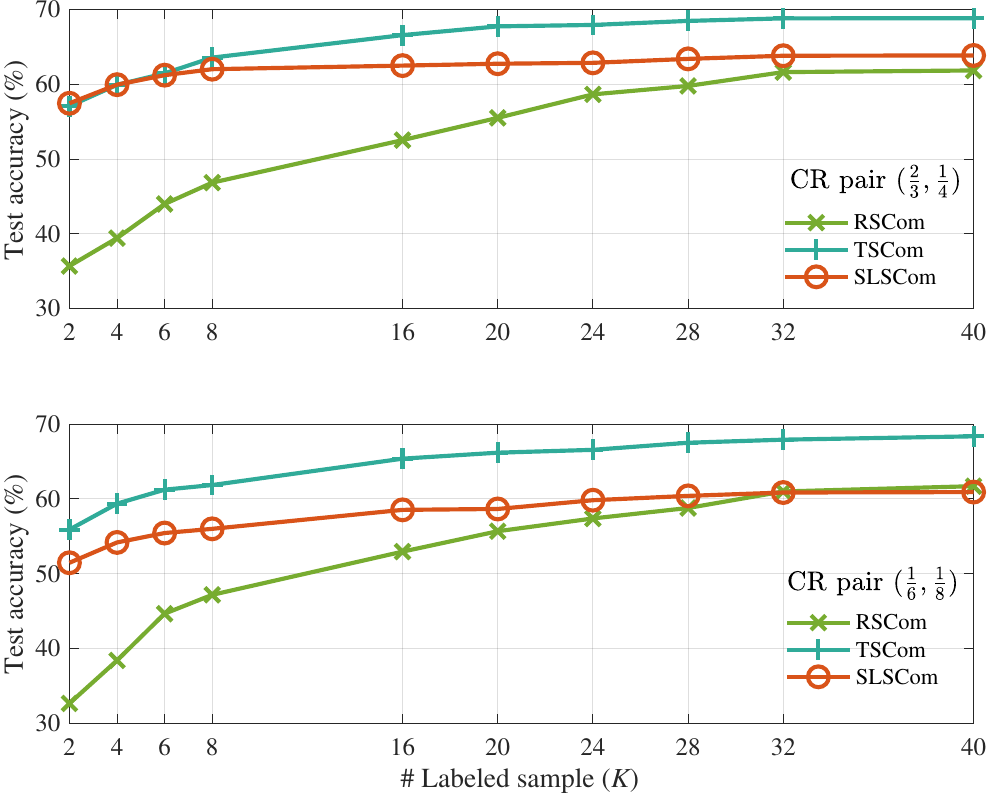}
    \caption{Test accuracy comparison on CIFAR10 in partial label missing scenarios.}
    \label{fig:pm}
    \vspace{-0.2cm}
	\end{figure} 
	
	\begin{figure}[t]
	    \centering
	    \includegraphics[width=0.65\linewidth]{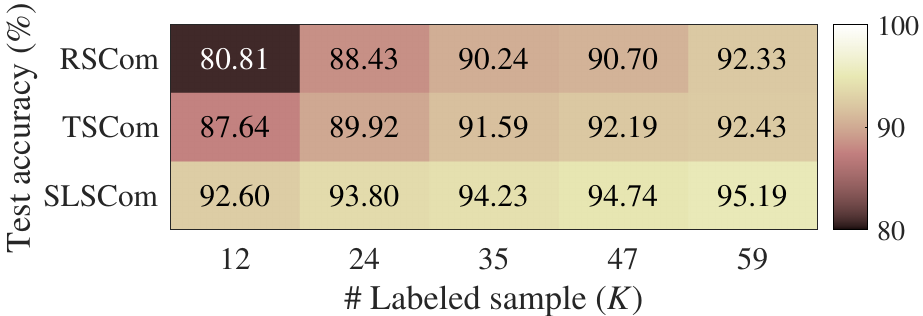}
	    \caption{Heat map of test accuracy on SVHN versus the number of labeled samples.}
	    \label{fig:svhn}
	    \vspace{-0.5cm}
	\end{figure}

	To validate the capability of the self-supervisedly learned semantic knowledge in dealing with data from unseen distributions, in addition to SLSCom w DS, we incorporate scenarios where partial labels are missing during the network training. The superior performance of SLSCom w DS, as seen in Fig.~\ref{cifar10_acc}, suggests the strong generalizability of the self-supervisedly learned semantic encoder on data from different distributions. The simulation settings in SLSCom w DS reflect practical scenarios where data surrounding edge devices may exhibit distributions different from task-specific data, regarding data content and label sets. Additionally, we purposefully exclude two categories of labeled samples in the partial label missing scenarios. As illustrated in Fig.~\ref{fig:pm}, methods with pre-trained models, including TSCom and SLSCom, demonstrate superior performance, particularly with a small set of labeled data. Due to being supervisedly pre-trained on the large-scale ImageNet dataset of $1000$ classes, TSCom containing rich knowledge achieves the highest test accuracy. Importantly, SLSCom is able to closely match TSC when using a deeper NN structure and few labeled samples. These validations help support the claim that the semantic knowledge acquired from unlabeled data can effectively handle data from unseen distributions.

	We compare the test accuracy under a different SNR level in Table~\ref{tab:acc_nlabel_4dB}, where all methods are trained and test at $-4$~dB. All approaches exhibit performance degradation, particularly in scenarios with few labeled samples, such as fewer than $8,000$ labeled samples. SLSCom shows the smallest decrease in performance among the three methods. This suggests that the proposed framework is less sensitive to adverse SNR conditions, showcasing its robustness against variations in SNR conditions.
	
	\rev{We also test the effectiveness of SLSCom on the larger-scale SVHN dataset, with the number of labeled samples ranging from $12, 000$ to $59, 000$.} The results, shown in \ref{fig:svhn}, indicate that SLSCom consistently outperforms both RSCom and TSCom, requiring only $1/6$ of the labeled samples. Additionally, the performance advantage of SLSCom is more pronounced in scenarios with few labeled samples, demonstrating its outstanding advantages in low-label applications in practice.	

\subsection{Performance versus SNR}
	In this subsection, we evaluate the robustness of the proposed framework, using the deeper NN structure, under different SNR levels as defined in \eqref{eq:snr}. 
	
	\begin{figure}[t]
		\centering
		\includegraphics[width=0.75\linewidth]{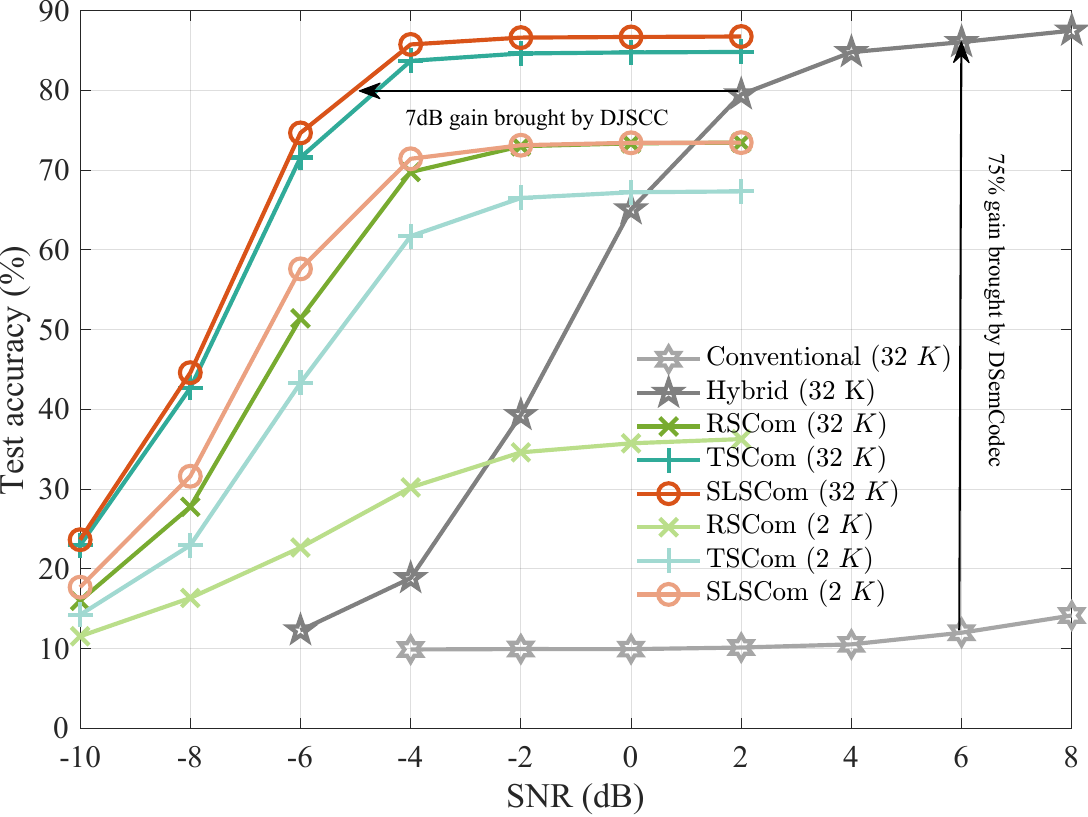}
		\caption{Test accuracy comparison on CIFAR10 versus SNRs.}
		\label{cifar_snr}
	\end{figure}
	\begin{figure}[t]
		\centering
		\includegraphics[width=0.75\linewidth]{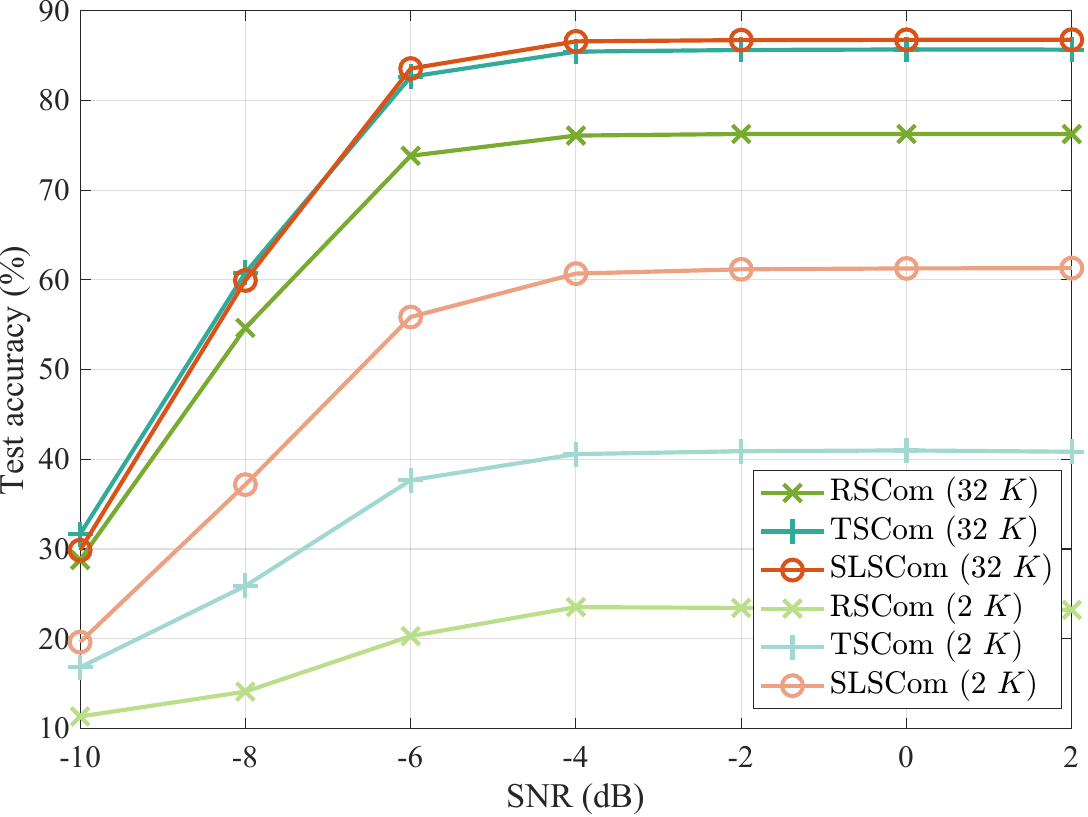}
		\caption{Test accuracy comparison on CIFAR10 versus SNRs with a training SNR of $-4$~dB.}
		\label{cifar_snr_-4}
	\end{figure}
	
	Fig.~\ref{cifar_snr} illustrates the classification performance across different test SNRs. The curves labeled as `$32\ K$' and `$2\ K$' represent methods using $32,000$ and $2,000$ labeled samples during the end-to-end supervised learning stage, respectively. SLSCom consistently outperforms other DL methods, demonstrating robustness against varying SNR conditions. Additionally, comparisons with conventional and hybrid digital coding baselines clearly identify the sources of performance improvements. Specifically, we observe a $75\%$ accuracy increase for the hybrid method over the conventional approach in high SNR regions, underscoring the effectiveness of the deep semantic encoder and decoder networks trained under the proposed framework. This highlights the robustness of the extracted semantic information against channel noise. Furthermore, SLSCom outperforms the hybrid method, achieving a $7$~dB improvement at $80\%$ accuracy. This performance gain underscores the advantages provided by the JSCC network and the benefits of joint training. Notably, SLSCom significantly reduces the required transmission volume, utilizing only $1/36$ and $1/24$ of the transmission symbols required by the conventional and hybrid baselines, respectively.
	
	Fig.~\ref{cifar_snr_-4} presents the performance comparisons under different test SNRs with training SNR $-4$~dB. SLSCom demonstrates superior performance, especially with limited labeled samples. By comparing Fig.~\ref{cifar_snr} and Fig.~\ref{cifar_snr_-4}, we have made the following observations. When there are sufficient labeled samples, e.g., $32,000$ samples, the training SNR significantly impacts performance. Specifically, methods trained at an SNR of $-4$~dB exhibit better performance in lower test SNR regimes compared to those trained at $0$~dB. However, when the number of labeled samples is limited, we observe a different trend with RSCom and TSCom. Despite being trained under low SNR conditions, these two methods show inferior performance in low test SNR scenarios compared to their performance when trained under higher SNR conditions. This suggests that RSCom and TSCom are more sensitive to the quantity of labeled samples and show a demonstrated dependency on it.

\begin{figure}[t]
	    \centering
	    \includegraphics[width=0.75\linewidth]{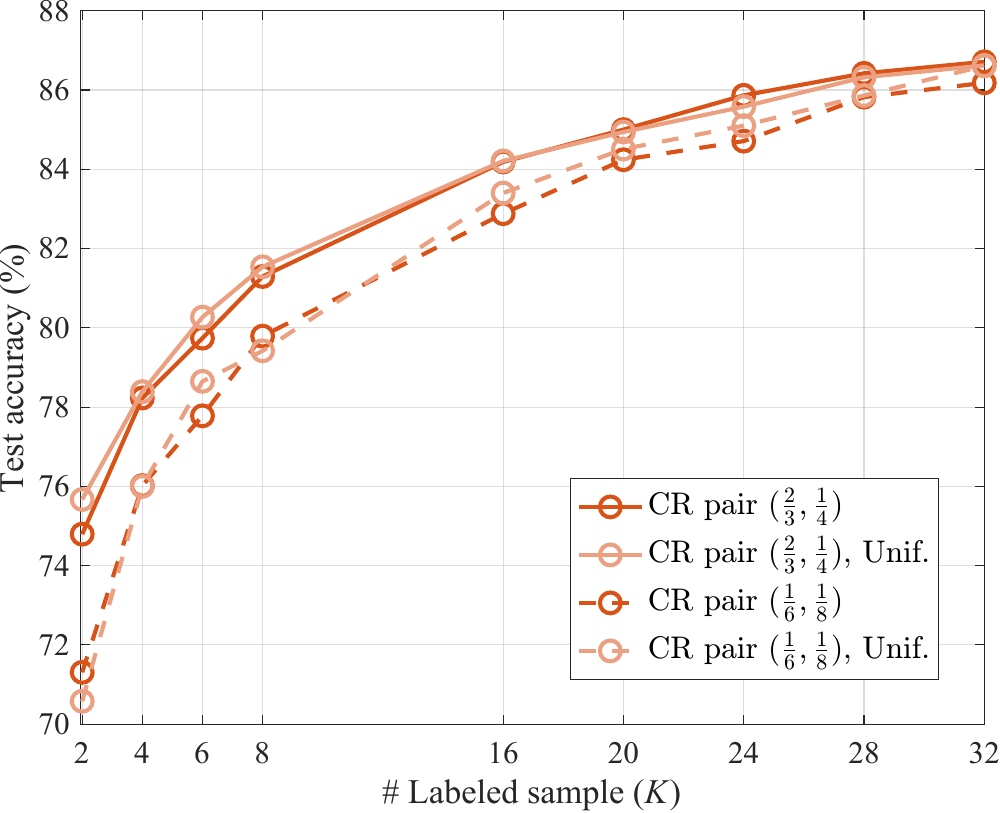}
	    \caption{Test accuracy comparison on CIFAR10 versus the number of labeled samples with different number distributions of labeled samples.}
	    \label{fig:avgdata}
	    \vspace{-0.2cm}
	\end{figure}
	

\subsection{Impact of the Number Distribution of Labeled Samples}
	
In this subsection, we clarify the distribution of the number of labeled samples across different classes and discuss the impact of this distribution on classification performance. In our simulations, the number of training samples per class varies due to the random splitting of the training and validation sets. This variation reflects real-world scenarios where labeled samples are often not evenly distributed among classes due to factors such as data collection bias and natural class imbalance.
Additionally, we consider a setup in which each class has an equal number of samples, providing a more balanced dataset. The results are detailed in Fig.~\ref{fig:avgdata}.
It is observed that SLSCom with a uniform number of samples per class, referred to as an `Unfi.' notation, performs slightly better than when there are uneven numbers of samples per class. We conjecture that having an equal number of samples per class facilitates balanced feature learning across all classes, which enhances the model’s generalizability and stability.

\begin{table*}[ht]
	\centering
	\caption{Ablation Study for Auxiliary Projection: Test Accuracy (\%) versus the Number of Labeled Samples on CIFAR10}
	\label{tab:woap}
	\resizebox{\linewidth}{!}{%
	\begin{threeparttable}
	\begin{tabular}{cc|c|c|c|c|c|c|c|c|c|c}
	\toprule
	\multicolumn{2}{c|}{\# Labeled sample $(K)$}  & $2$ & $4$ & $6$ & $8$ & $16$ & $20$ & $24$ & $28$ & $32$ & $40$ \\ \hline
	\multicolumn{1}{l|}{\multirow{2}{*}{SNR~$0$~dB}} &SLSCom& $73.23$ & $78.15$ & $79.17$ & $80.37$ & $83.87$ & $85.08$ & $85.73$ & $86.10$ & $86.61$ & $87.60$ \\ \cline{2-12}  
	\multicolumn{1}{l|}{} & SLSCom w/o AP & $74.80$ & $77.59$ & $79.74$ & $81.30$ & $84.19$ & $85.00$ & $85.87$ & $86.42$ & $86.71$ & $87.86$ \\ \hline 
	\multicolumn{1}{l|}{\multirow{2}{*}{SNR~$-4$~dB}} &SLSCom& $60.69$ & $74.72$ & $78.91$ & $79.17$ & $ 83.18$ & $84.24$ & $85.06$ & $85.73$ & $ 86.32$ & $87.30$ \\ \cline{2-12} 
	\multicolumn{1}{l|}{} & SLSCom w/o AP & $46.64$ & $69.94$ & $77.28$ & $80.16$ & $84.18$ & $84.61$ & $84.83$ & $85.66$ & $86.56$ & $87.59$ \\ \bottomrule
	\end{tabular}%
	\end{threeparttable}
	}
	\vspace{-5pt}
	\end{table*}

\begin{table*}[t]
	\centering
	\caption{Ablation Study for Auxiliary Projection: Test Accuracy (\%) versus Test SNR (dB) on CIFAR10 with Training SNR $-4$ dB}
	\label{tab:woapsnr}
	\resizebox{0.8\linewidth}{!}{%
	\begin{tabular}{cc|c|c|c|c|c|c|c}
	\toprule
	\multicolumn{2}{c|}{Test SNR (dB)} & $-10$ & $-8$ & $-6$ & $-4$ & $-2$ & $0$ & $2$ \\ \hline
	\multicolumn{1}{c|}{\multirow{2}{*}{$32~K$ Labeled samples}} &SLSCom& $29.72$ & $59.59$ & $83.26$ & $86.32$ & $86.45$ & $86.46$ & $86.44$ \\ \cline{2-9}
	\cline{2-9}
	\multicolumn{1}{c|}{} &SLSCom w/o AP& $29.58$ & $58.93$ & $83.55$ & $86.56$ & $86.72$ & $86.75$ & $86.76$ \\ \hline
	\multicolumn{1}{c|}{\multirow{2}{*}{$2~K$ Labeled samples}} & SLSCom & $19.62$ & $ 37.17$ & $55.86$ & $60.69$ & $61.17$ & $61.27$ & $61.30$ \\ \cline{2-9}
	\multicolumn{1}{c|}{} &SLSCom w/o AP & $8.70$ & $24.36$ & $40.81$ & $46.64$ & $47.24$ & $47.34$ & $47.38$ \\ \bottomrule 
	\end{tabular}%
	}
	\vspace{-5mm}
	\end{table*}
		
	\begin{figure}[t]
	    \centering
	    \includegraphics[width=0.72\linewidth]{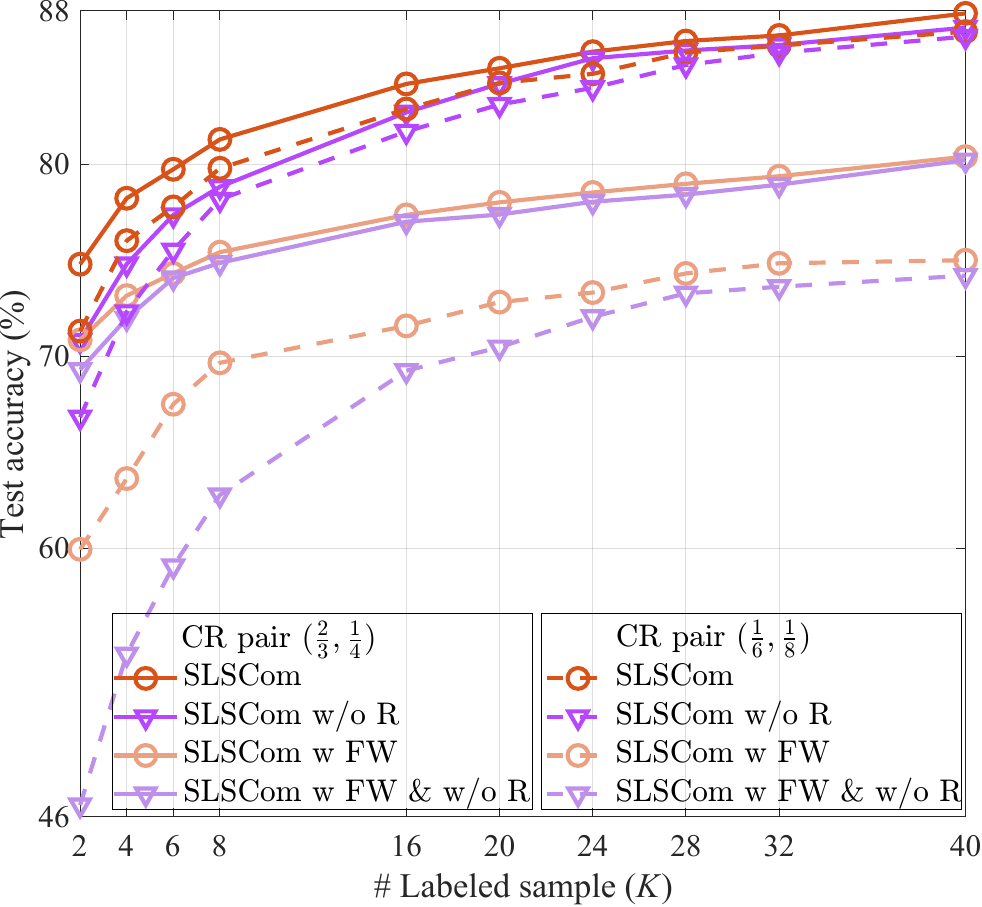}
	    \caption{Ablation study for reconstruction pretext task: test accuracy comparison versus the number of labeled samples.}
	    \label{fig:woR}
		\vspace{-5pt}
    \end{figure}

\subsection{Ablation Study}  
In this subsection, we conduct three ablation studies on the proposed learning framework to verify the necessity of the reconstruction pretext task, as well as to assess the impact of auxiliary projection (AP) and color-related transformations on inference performance. 
  
  \subsubsection{For Reconstruction Pretext Task} We compare the performance between SLSCom and its ablation of reconstruction pretext task in Fig.~\ref{fig:woR}.
  It is found that SLSCom slightly outperforms SLSCom w/o R when using the deeper NN structure. With a lighter NN, the performance gap between the two methods widens, especially freezing the parameters of the pre-trained semantic encoder, i.e., the SLSCom w FW method. This suggests that the lighter network relies more heavily on the guidance provided by the reconstruction pretext task. Recall that the difference between the two NN structures lies in the dimensionality of both semantic features and channel input symbols. The deeper network produces semantic features and channel input symbols of higher dimensionality, enabling it to better tolerate the presence of task-irrelevant information. In contrast, the lighter network, constrained by its reduced dimensionality, must ensure that these features contain as much task-relevant information as possible. In such situations, discarding task-irrelevant information becomes crucial to maximize the effectiveness of the limited feature dimensions.

	In conclusion, compared to the deeper network, the lighter network benefits more significantly from the inclusion of the reconstruction pretext task, as it enhances their ability to extract and transmit task-relevant information effectively within their limited capacity.

  \subsubsection{For Auxiliary Projection} Table~\ref{tab:woap} and Table~\ref{tab:woapsnr} show the test accuracy with and without AP across varying numbers of labeled samples and SNR conditions. 
As shown in Table~\ref{tab:woap}, when trained with a sufficient number of labeled samples at a high SNR, SLSCom achieves comparable inference performance to that of SLSCom without AP. This suggests that, in such favorable conditions, AP may not be essential and can be removed without significantly affecting the system performance. 
Conversely, SLSCom demonstrates substantial performance improvement over SLSCom without AP when trained at a low SNR and with a limited number of labeled samples, specifically at $-4$ dB and fewer than $6,000$ samples.	
	
We subsequently investigate performance comparisons for methods trained under low SNR conditions, as shown in Table~\ref{tab:woapsnr}. It is observed that SLSCom significantly outperforms SLSCom without it when trained with few labeled samples. These performance gains are noticeable across different test SNRs, highlighting the ability of AP to enhance the model's robustness and effectiveness in challenging environments. In such cases, the AP layer proves essential for improving feature discrimination and overall task performance.

In conclusion, while the auxiliary projection layer may be omitted in high SNR and data-rich scenarios to reduce computational complexity without compromising performance, it is beneficial and necessary under low SNR and limited labeled data conditions to boost the model's robustness and accuracy.

\begin{table}[t]
		\centering
		\caption{Ablation Study for Color-related Transformations: Test Accuracy~(\%) versus the Number of Labeled Samples on Flower Recognition}
		\label{tab:wocolor}
		\begin{tabular}{c|c|c|c}
		\toprule
		\# Labeled sample (K) & $140$ & $1400$ & $2250$ \\ \hline
		SLSCom & $23.50$ & $75.50$ & $77.25$ \\ \hline
		SLSCom w/o color transf. & $20.00$ & $65.00$ & $70.38$ \\ \bottomrule
		\end{tabular}%
	\vspace{-15pt}
	\end{table}  
  \subsubsection{For Color-related Transformations}
  
To better assess the impact of color-related transformations on task inference performance, we choose the flower recognition dataset, which contains flower images with rich colors and clear details, similar to those in ImageNet. The performance comparison using the deeper NN structure is shown in Table~\ref{tab:wocolor}. The color-related transformations, including ColorJitter and grayscale, positively affect task inference performance. The results align with the goal of applying these color-related transformations, which is to help improve the model's generalization by learning robust features that are invariant to certain changes.

\section{Conclusion}
	In this paper, we propose an efficient learning approach by utilizing a self-supervised learning framework for task-oriented semantic communication, which improves task inference performance under wireless multipath channels, especially in challenging scenarios such as low SNR conditions and scarce labeled training samples.
	We develop a task-relevant semantic encoder using unlabeled data by introducing self-supervision and devising classification and reconstruction pretext tasks. This approach provides a practical and effective solution to the IB problem, which balances the tradeoff between the informativeness of the extracted semantic information and task inference performance. The task-relevant semantic extraction capability is further strengthened through efficient end-to-end learning, which is jointly conducted with the JSCC and semantic decoder networks. Extensive simulation results demonstrate that the proposed approach outperforms digital coding, training from scratch, and transferring pre-trained model methods. Moreover, we validate its robust generalizability under complex settings, including data distribution shifts and partial label missing.
	The proposed learning approach holds potential for extension to multi-device settings, as these environments can provide significant benefits in terms of data volume and diversity, which in turn can improve the quality of the learned contrastive features.
	This extension presents several challenges, including data heterogeneity, communication delays between devices, and variations in data characteristics across devices. Future research will focus on addressing these challenges and evaluating the effectiveness of the proposed approach in multi-device settings. 


%

%
%



\ifCLASSOPTIONcaptionsoff
  \newpage
\fi



\bibliographystyle{IEEEtran}
\bibliography{myrefers.bib}
%



\end{document}